\pdfoutput=1

\documentclass[11pt]{article}

\usepackage{acl}

\usepackage{times}
\usepackage{latexsym}

\usepackage[T1]{fontenc}

\usepackage[utf8]{inputenc}

\usepackage{microtype}

\usepackage{inconsolata}

\usepackage{hyperref}
\usepackage{url}
\usepackage{xcolor}
\usepackage{soul}
\usepackage{amsmath}
\usepackage{mathtools}
\usepackage{bbm}
\usepackage{multirow}
\usepackage{multicol}
\usepackage{makecell}
\usepackage{booktabs}
\usepackage{xfrac}
\usepackage[export]{adjustbox}
\usepackage{lipsum}
\usepackage[shortlabels]{enumitem}
\usepackage{amssymb}
\usepackage{listings}
\usepackage[linesnumbered,ruled,vlined]{algorithm2e}

\SetCommentSty{mycommfont}
\usepackage{wrapfig}
\usepackage{graphicx}
\usepackage{caption}
\usepackage{subcaption}
\usepackage{comment}
\usepackage{multicol}
\usepackage{amsthm}
\usepackage{framed} 
\usepackage[symbol]{footmisc}
\definecolor{dkgreen}{rgb}{0,0.6,0}
\definecolor{gray}{rgb}{0.5,0.5,0.5}
\definecolor{mauve}{rgb}{0.58,0,0.82}
\newcommand{\hlc}[2][yellow]{{%
    \colorlet{foo}{#1}%
    \sethlcolor{foo}\hl{#2}}%
}
\newcommand{\edit}[1]{\textcolor{black}{#1}}

\makeatletter
\newcommand{\distas}[1]{\mathbin{\overset{#1}{\kern\z@\sim}}}%
\newsavebox{\mybox}\newsavebox{\mysim}
\newcommand{\bigO}{\mathcal{O}}
\makeatother

\newcommand{\cs}{\texttt{Cyclic-Shift}}
\newcommand{\mh}{\texttt{Message-Hash}}
\newcommand{\ours}{\texttt{MPAC}}
\newcommand{\expnumber}[2]{{#1}\mathrm{e}{#2}}

\makeatletter
\usepackage{cleveref}
\crefformat{section}{\S#2#1#3}
\crefformat{subsection}{\S#2#1#3}
\crefformat{subsubsection}{\S#2#1#3}
\title{Advancing Beyond Identification: \\Multi-bit Watermark for Large Language Models via Position Allocation}

\author{KiYoon Yoo\textsuperscript{1}\hspace{.2cm} 
        Wonhyuk Ahn\textsuperscript{2}\hspace{.2cm}
        Nojun Kwak\textsuperscript{1}\thanks{Corresponding author}\hspace{.2cm}
        \\
\textsuperscript{1}{Seoul National University}\hspace{.2cm}
\textsuperscript{2}{Webtoon AI} \\
\texttt{\{961230,nojunk\}@snu.ac.kr} \hspace{.2cm} \texttt{whahnize@gmail.com}
}

\begin{document}
\maketitle
\begin{abstract}
    \label{sec:abstract}
    We show the viability of tackling misuses of large language models beyond the identification of machine-generated text. While existing zero-bit watermark methods focus on detection only, some malicious misuses demand tracing the adversary user for counteracting them. To address this, we propose Multi-bit Watermark via Position Allocation, embedding traceable multi-bit information during language model generation. Through allocating tokens onto different parts of the messages, we embed longer messages in high corruption settings without added latency. By independently embedding sub-units of messages, the proposed method outperforms the existing works in terms of robustness and latency. Leveraging the benefits of zero-bit watermarking~\citep{kirchenbauer2023watermark}, our method enables robust extraction of the watermark without any model access, embedding and extraction of long messages ($\geq$ 32-bit) without finetuning, and maintaining text quality, while allowing zero-bit detection all at the same time. Code is released here: \href{https://github.com/bangawayoo/mb-lm-watermarking}{https://github.com/bangawayoo/mb-lm-watermarking}.
\end{abstract}

\section{Introduction}
    \label{sec:intro}
 How can we take a step further from merely identifying machine-generated text to proactively tackling misuses of large language models? The emergence of \edit{human-like language models and their easily accessible nature via web interface and APIs have garnered unprecedented attention from the public and academia~\citep{chatgpt-users}.  The ability to follow complex instructions has boosted the productivity of various tasks such as programming, creative writing, and more.  However, there have been increasing concerns about exploiting such language models to automate malicious activities such as spreading disinformation.} This has necessitated the development of various methods to detect machine-generated texts through techniques such as zero-shot detection, supervised training, watermarking, and more \citep{mitchell2023detectgpt, wang2023m4, kirchenbauer2023watermark, krishna2023paraphrasing}. 
These endeavors focus on the crucial task of \textit{identifying} machine-generated content, which serves as a pivotal step in mitigating the potential harm caused by such text. 

However, when it comes to more pernicious misuses of large language models, such as the dissemination of misinformation  and war propaganda on social media platforms for political or financial gains~\citep{Badway2018, pierri2023propaganda, amazon-reviews}, the stakes are considerably higher, potentially leading to the erosion of social trust~\citep{valenzuela2022downward}. 
In such circumstances, merely identifying the machine-generated text may not suffice for the language model providers. Instead, the ability to trace back to the adversary user responsible for generating the content becomes pivotal in counteracting such misuses. By doing so, the API providers can take a precursory measure to ban these users from their systems and allow media and social platforms, along with API providers, to collaborate with law enforcement authorities and take more decisive actions. All in all,  watermarking the user information (or part thereof) can hold the adversary user accountable for potential harms facilitated through language model APIs without having to store user queries~\citep{krishna2023paraphrasing}, which would be prohibitively expensive and concern ordinary users who value privacy. 

\begin{figure*}[t]
    \centering
    \includegraphics[width=.9\textwidth]{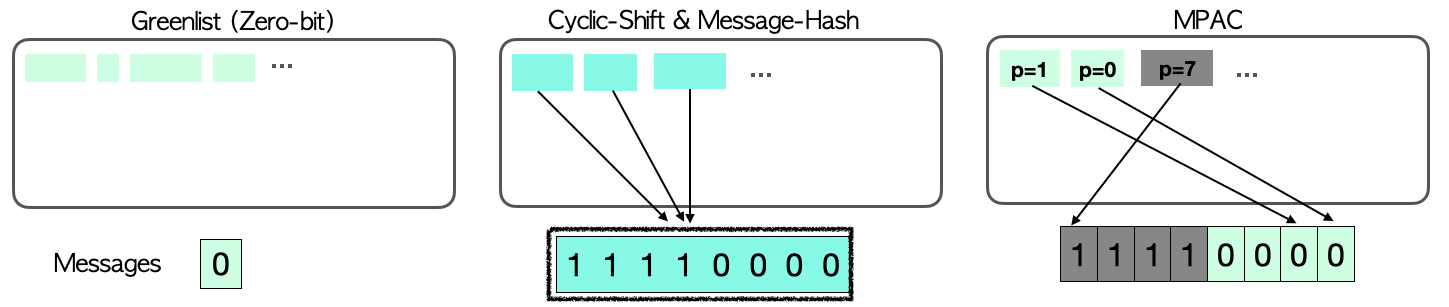}
    \caption{Comparison of how messages are encoded for zero-bit watermarking~\citep{kirchenbauer2023watermark}, recent multi-bit methods, and our proposed method \texttt{MPAC}. For \texttt{MPAC}, the number inside a token (e.g. $\boxed{p=1}$) denotes the allocated position.}
    \vspace{-3mm}
    \label{fig:overview-comparison}
\end{figure*}

All this can be achieved by embedding multi-bit information. Recent works~\citep{fernandez2023stable, wang2023towards} have achieved this by providing a distinct signal for each multi-bit message.  While this is effective in low bit-width and low noise settings, maintaining the integrity of the watermark becomes increasingly difficult as the bit-width increases due to the exponential number of possible messages. This is further aggravated in the presence of higher noise. In addition, having to consider all the possible messages also follows with the side effect of increased latency during encoding and/or decoding phase, the former of which degrading the end user experience.

As opposed to this, our proposed method \textbf{M}ulti-bit watermark via \textbf{P}osition \textbf{A}llo\textbf{c}ation (\texttt{MPAC}) first allocates each token pseudo-randomly onto a sub-unit of the message to be embedded (Fig. \ref{fig:overview-comparison}). The allocation of tokens onto different parts of the messages allows the embedding of longer messages without added generation latency and fares well in high corruption settings. Then the message content at the allocated position determines the state to encode using a zero-bit watermarking scheme. For instance, when following the zero-bit watermarking scheme of \citet{kirchenbauer2023watermark}, the message content decides which token subsets are biased. To increase load capacity, we can further partition the vocabulary into multiple ``colored'' lists instead of a single green list, effectively encoding multiple states for every token. Our experiments show our method improves upon the runner-up baseline in terms of robustness of the watermark $\geq$20\% in high-noise setting for 16-bit and 24-bit messages. 

\vspace{-3mm}

\section{Related Works}
    \label{sec:related}
    Watermarking has been studied in various types of multimedia such as image~\citep{potdar2005survey}, video~\citep{asikuzzaman2017overview}, audio~\citep{hua2016twenty}, and natural language~\citep{topkara2005natural}.
Following previous works~\citep{zhu2018hidden, luo2020distortion}, we use the term watermarking to denote embedding information into natural language in a manner that is robust against possible attacks given a watermarked text -- in our case, this is the output generated by a language model given the prompt. This differs from steganography~\citep{cheddad2010digital, fang2017generating, ziegler2019neural, de2022perfectly}, which focuses more on the undetectability of a secret message that is embedded in the multimedia rather than robustness. \edit{For instance, \citet{ziegler2019neural} sequentially encodes information via arithmetic coding every token. Naively applying this deterministic encoding scheme makes the watermark extremely fragile to simple corruptions as shown in Appendix Fig.~\ref{fig:fragile-coding}}.

Recently, methods relying on neural networks have shown progress in natural language watermarking, outperforming traditional methods that rely on rule-based watermarks~\citep{topkara2006hiding, topkara2006natural, atallah2001natural}. \citet{abdelnabi2021adversarial} proposed an end-to-end framework where a decoder network predicts the encoded message. \citet{yang2022tracing} improved upon the quality of the watermarked text by using an algorithmic approach. Building upon this, \citet{yoo2023robust} focused on robustness and capacity, outperforming previous works on both aspects. However, since the proposed method works at the sentence-level, any addition or removal of a sentence will fail to extract the watermark. Moreover, these works cannot distinguish non-watermarked texts, making them unsuitable for distinguishing between machine text and human text.

Meanwhile, directly watermarking language models in a zero-bit manner during token generation has emerged as a promising approach for distinguishing language model outputs from human text~\citep{kirchenbauer2023watermark, openai-watermark} while achieving robustness against realistic attacks \citep{kirchenbauer2023reliability}. Several works have improved upon~\citet{kirchenbauer2023watermark}, e.g., in low entropy generation tasks such as code generation~\citep{lee2023wrote}, undetectability of the watermark~\citep{christ2023undetectable}, and its robustness~\citep{munyer2023deeptextmark}. We focus on extending the prior work for a more proactive counteraction towards identifying malicious users of language models by embedding \textit{any} information while maintaining the key advantages.

Concurrent to our work, \citet{fernandez2023three} and \citet{wang2023towards} use the entire message to create a signal unique to each message. Crucially, both works use the entire message content directly during embedding as input to the random seed generator, which leads to key differences in terms of robustness and latency. We further discuss their methodology in comparison with ours in the next section. Aside from this, \citet{wang2023towards} further utilize a proxy language model to enhance text quality. 

To give a rough estimate of the required message length for encoding a user ID,  consider the POSIX \citep{posix} standard used when creating usernames in operating systems. 65 characters ($\sim$7 bits) are permitted by POSIX, meaning at least 35 bits are required to encode a username of 5 characters. Accordingly, works in image watermarking embeds messages easily over 32-bits~\citep{zhu2018hidden, zhao2023recipe, fernandez2023stable}. Our method makes this feasible by encoding each bit position independently.

\section{Method}
    \label{sec:method}

We outline the multi-bit watermark protocol:
\begin{enumerate}
\item A user sends a prompt $X$ to the language model provider.

\item Using the message encoding function $\mathcal{E}$ , the language model provider generates watermarked text $Y$ embedded with a multi-bit information. The message contains user-specific meta-data that can aid tracing back to the user (e.g. timestamp, location, ID). 

\item The user publishes the text $\tilde Y$, which may be edited from the original watermarked text.

\item If the published text is deemed unsafe or malicious, the detector inspects $\tilde Y$  (i) to determine whether the watermark is present (zero-bit detection) and (ii) decode the multi-bit message to take further measure.
\end{enumerate}

\subsection{Zero-bit Watermarking}\label{subsec:zwb}
Throughout the paper, we focus on applying our multi-bit framework using the zero-bit zero-bit watermarking scheme introduced in \citet{kirchenbauer2023watermark}.\footnote{Extension to other schemes are in Appendix \ref{appendix:block-allocation}.} 
As a preliminary, we briefly review the scheme. An auto-regressive language model $p(y|x)$ predicts the probability distribution over the next token $\Delta(\mathcal{V})$ given arbitrary length prefix tokens where $\mathcal{V}$ is the vocabulary. A zero-bit watermark is embedded by biasing the language model to output a certain subset of tokens. That is, the message encoding function $\mathcal{E}:\Delta(\mathcal{V}) \rightarrow \Delta(\mathcal{V})$ generates another probability distribution that alters the original distribution of $p(y|x)$.

For \citet{kirchenbauer2023watermark}, the message encoding function pseudo-randomly chooses a subset of tokens at each token step $t$ to form a green list $\mathcal{G}_t$. The logit scores  $l_t \in \mathbb{R}^{|\mathcal{V}|}$ are modified towards selecting the green-listed tokens in favor of the other tokens by adding a bias term $\delta$ to the logits in $\mathcal{G}_t$. Instead of fixing the greenlist using rule-based heuristics such as spelling or synonym variations \citep{he2022protecting}, the greenlist is selected pseudo-randomly at each time step to minimize a noticeable shift in text distributions. At each time step, a seed $s$ is outputted depending on the previous $h$ tokens using a pseudo-random function $f: \mathbb{N}^h \rightarrow \mathbb{N}$, and $s$ is used to sample $\mathcal{G}_t$ from $\mathcal{V}$. 

We dub this message encoding function as $\texttt{Greenlist}$. Given $t-1$ prefix tokens $X_{1:t-1}$, and pseudo-random function $f$, the $t^\text{th}$ token is generated by 

\FrameSep1pt{
\begin{framed}
\noindent\textbf{{\texttt{Greenlist}}}
\begin{enumerate}[noitemsep]
    \item Compute hash of tokens $s=f(X_{t-h:t-1})$.
    \item Permute vocabulary $\mathcal{V}_t$ using $s$ as seed for a random number generator (RNG).
    \item Let $\mathcal{G}_t$ be the first $\gamma |\mathcal{V}|$ tokens from $\mathcal{V}_t$
    \item Add $\delta$ to token logits in $\mathcal{G}_t$.
\end{enumerate}
\end{framed}}

\noindent\textbf{Decoding.}
To determine the presence of the watermark, the detector inspects the ratio of the green-listed token using the same pesudo-random function $f$. A watermarked text will ideally have a high ratio of green tokens. Without the knowledge of the greenlist (null hypothesis), the number of tokens in the greenlist ($g$) follows a binomial distribution. \citep{kirchenbauer2023watermark} used the normal approximation to the binomial distribution to compute the $z$-statistics for a text with $T$ tokens: $z = \frac{g - \gamma T}{\sqrt{\gamma(1-\gamma)T}}$.

\subsection{MPAC: Extending to Multi-bit Watermark}
\label{subsec:multibit}
The objective of multi-bit watermarking is to embed and extract a message $\mathbf{m} \in \Sigma^b$ where $\Sigma$ denotes the $r$-ary possible strings, or more commonly referred to as the alphabet. For a binary message, $\Sigma=\{0,1\}$.  We let $p \in \{0, \dots, b-1\}$ denote the position of the message and $\mathbf{m}[p] \in \{0, \dots, r-1\}$ the message content at position $p$. Hereafter, we use $[a]$ to denote the integer set $\{0, \dots, a-1\}$. 

Our proposed method \textbf{M}ulti-bit watermarking via \textbf{P}osition \textbf{A}llo\textbf{c}ation (\ours) works by allocating the tokens to message positions. First, notice that zero-bit watermarking can be viewed as watermarking a single bit of information stating the existence of a watermark. In essence, each token generated by the language model is a signal that reinforces the watermark. 

\begin{figure}[t]
    \centering
    \includegraphics[width=.48\textwidth]{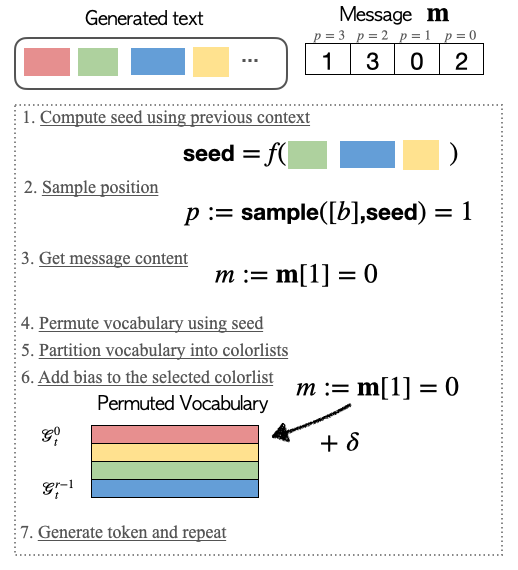}
    \caption{An overview of our method MPAC. See \cref{subsec:multibit} for details.}
    \label{fig:overview}
\vspace{-4mm}
\end{figure}

Our message encoding function $\mathcal{E}:\Sigma^{b} \times \Delta(\mathcal{V}) \rightarrow \Delta(\mathcal{V})$ alters the probability distribution dependent on the message. We first assign a position $p$  using a random number generator seeded with $s$. Then the message content $m=\mathbf{m}[p] \in [r]$ is encoded by permuting $\mathcal{V}$ and favoring the $m^{\text{th}}$ subset. Our message encoding function is extremely easy to implement over the $\texttt{Greenlist}$ scheme. We highlight the steps in colors that are specific to ours. An illustrative flow diagram is shown in Fig. \ref{fig:overview}.

\FrameSep1pt{
\begin{framed}
\noindent\textbf{\texttt{MPAC}}
\begin{enumerate}[noitemsep]
    \item Compute $s=f(X_{t-h:t-1})$.
    \item \hlc[cyan!40]{$p \leftarrow \texttt{sample}([b])$ using $s$ as seed.} 
    \item \hlc[cyan!40]{$m \leftarrow \mathbf{m}[p]$} 
    \item Permute vocabulary $\mathcal{V}_t$ using $s$ as seed.
    \item \hlc[cyan!40]{Partition $\mathcal{V}_t=[\mathcal{C}_t^{0}, \cdots , \mathcal{C}_t^{r-1}]$ discarding remainders if any.}
    \item Add $\delta$ to token logits in $\mathcal{C}_t^{m}$.
\end{enumerate}
\end{framed}}

\noindent\textbf{Colorlisting}
In Step 5, $r$ is the number of available partitions. The number of vocabulary partitions is determined by the greenlist proportion $\gamma$, i.e. $r = \lfloor \frac{1}{\gamma} \rfloor$.  When $r>2$, we can further increase the load capacity by taking advantage of all the `colored' lists (hence, the notation $\mathcal{C}$), instead of only using the greenlist. Given a binary message of length $b$, the message is convereted to radix $r$ attaining $\mathbf{m}_r \in [r]^{\tilde b}$ where $\tilde b=\lceil \frac{b}{\log_2{r}} \rceil $. In Fig. \ref{fig:overview}, we illustrate the case of $r=4$ and $b=8$, where the 8-bit message is converted into radix $4$, resulting in an effective message length of $\tilde b=4$. 
When $\tilde b \neq b,$ we sample the position from $\tilde b$.

At each token generation, the message content at the assigned position $p$ determines which colorlist to add $\delta$ to. If the message content is `0', the tokens from the first list (red in Fig. \ref{fig:overview}) are favored. Note that zero-bit watermarking can be seen as a special case of embedding the same single bit message ($b=1, \mathbf{m}=0$).

\noindent \textbf{Message Decoding}
Given a watermarked language model output, we determine the position and which colorlist each token is from and increment the number of tokens in the colored lists. For instance, for the $t^\text{th}$ token with message position $p=i$ and the $j^{\text{th}}$ colorlist $\mathcal{C}_t^j$, we increment the counter $\mathbf{W}[i][j]$ where $\mathbf{W}\in\mathbb{R}^{\tilde b \times r}$. After computing this on the entire text segment, we predict the message content by taking the colorlist with the most tokens for each position. A Pythonic algorithm is shown in Algorithm \ref{alg:extraction}.

\SetKwInput{KwInput}{Input}              
\SetKwInput{KwOutput}{Output}  
\maketitle
\SetAlgoNoLine
    \begin{algorithm*}[t]
    \DontPrintSemicolon
        \KwInput{Text $X_{1:T}$, context width $h$, effective message length $\tilde b$, counter $\mathbf{W}\in\mathbb{R}^{\tilde b \times r}$}
        \KwOutput{Predicted message $\hat{\textbf{m}}$, number of colorlisted tokens $w$}
        \begin{multicols}{2}
        \tcc{Initialize counter}
        {$\textbf{W}[p][m] = 0 \text{   } \forall p,m$}\; 
        \tcc{Count tokens in colorlists}
        \For{$t \texttt{ in } [h+1, T]$}
        {
            $s=f(X_{t-h:t-1})$\;
            $p=\texttt{sample}([\tilde b])$ using $s$ as seed\;
            $\mathcal{V}_t=\texttt{permute}(\mathcal{V}_t$) using $s$ as seed\;
            \For{$m \texttt{ in }[r]$}
            {
                \If{$X_t \in \mathcal{G}_t^{m}$}
                    {
                    $\textbf{W}[p][m]$ += 1 \;
                    continue\;
                    }
            }
            
        }
        \tcc{Predict message}
        $\hat{\textbf{m}}_r = \text{`` "}$\; 
        $w=0$\;
        \For{$p \texttt{ in } [\tilde b]$}
        {
            $w$ += $\texttt{max}(\textbf{W}[p][:])$\;
            $\hat{m}=\texttt{argmax}(\mathbf{W}[p][:])$\;
            $\hat{\textbf{m}}_r$ += $\texttt{str}(\hat{m}$) \; 
        }
        Get bit message $\hat{\textbf{m}}$ by converting $\hat{\textbf{m}}_r$ \;
        \Return $\hat{\textbf{m}}$, $w$
    \end{multicols}
\caption{Message Decoding}
\label{alg:extraction}
\end{algorithm*}

\subsection{Detecting Machine Text}\label{subsec:detection}

While we can use MPAC to decode the multi-bit watermark in conjunction with another detection mechanism, MPAC alone can detect human text from watermarked text just like zero-bit watermarking. To distinguish between a watermarked text and a non-watermarked (human-written) text, we count the number of tokens assigned to the predicted message. This corresponds to $w$  in Line 12 of Algorithm ~\ref{alg:extraction}. We model the number of tokens in the argmax colorlist of position $i$ as a random variable $C_i \distas{H_0} \text{Binomial}(T_i, \gamma$) where $T_i$ is the number of tokens assigned to position $i$. 
As $C_0, \dots, C_{b-1}$ are independent for a fixed set of trials ($T_i, \dots, T_{b-1}$) and have the same success probability parameter, the sum of these is a binomial random variable as well:
\begin{equation}\label{eq:zero-bit-detection}
    C = C_0 + \cdots + C_{b-1} \distas{H_0} \text{Binomial}(T, \gamma)
\end{equation}
where $T=T_{0}+\cdots+T_{b-1}$. This reduces to the same random variable used in zero-bit watermarking and we can compute the z-statistics. More discussions regarding the details of the z-statistic and other possible statistics are outlined in Appendix \ref{appendix:detection-analysis}.

\subsection{Comparison to Other Works}
The message encoding function of existing works use the entire message $\mathbf{m}$. After permuting $\mathcal{V}_t$, \citet{fernandez2023three} cyclically shift the vocabulary $m_{\text{10}}$ times where $m_{\text{10}}$ is the radix-10 form of $\textbf{m}$. This modifies Step 2 of \texttt{Greenlist}. \citet{wang2023towards} hashes $\textbf{m}$ to attain a seed $s'$ to permute the vocabulary along with the seed attained from prefix tokens, modifying Step 1. 

\FrameSep1pt{
\begin{framed}
\noindent\textbf{\texttt{Cyclic-Shift}}
\begin{itemize}
    \item [2'.] Permute $\mathcal{V}_t$ using $s$ as seed. Then, cyclic shift $m_{\text{10}}$ times.
\end{itemize}
\textbf{\texttt{Message-Hash}}
\begin{itemize}
    \item [1'.] $s'\leftarrow$\texttt{Hash}~($s$+\texttt{Hash}~($m_\text{10}$))
\end{itemize}
\end{framed}}

Using the entire message leads to two key characteristics that diverge from ours. First, the hamming distance between two messages is not necessarily preserved after applying the encoding function. As an example, consider \texttt{Message-Hash}. Using the final seed $s'$ created from $\textbf{m}=0000$ does not guarantee an output from the RNG that is any closer to that of $\textbf{m}=0001$ (hamming distance of 1) as it is to $\textbf{m}=1111$ (hamming distance of 4). This leads to an all-or-nothing behavior where either the entire message is extracted without error or is a completely random message. In the presence of high corruption, which reflects the real-world case, we show this behavior is not desirable as it lacks enough signal to correctly predict the message. 

In addition,  the exponential number of messages ($\bigO(2^b)$)  should be considered during message decoding to find the optimal message, which renders decoding of long messages ($\geq 32$-bit) computation-heavy\footnote{See Section 7.5 of \citealp{wang2023towards}.
} . For \citealt{fernandez2023three}, the bit-width affects the \textit{encoding phase} due to the cyclic shift operation, which is more problematic as it affects the end users. MPAC encodes and decodes each bit position of the message independently, which brings a negligible increase in the computation as the message length is increased.  

The simplicity of our multi-bit watermark scheme via position allocation makes it easy to apply it on top of other methods. For example, using the position allocation scheme, we can decompose the multi-bit message into blocks and hierarchically embed them using the message encoding scheme of \citet{fernandez2023three}. Details are in Appendix \ref{appendix:list-decoding}. In addition, the message encoding function of \ours~can be generalized to other zero-bit watermark approaches that uses the exponential minimum sampling approach~\citep{openai-watermark, kuditipudi2023robust}. The scheme is provided in Appendix \ref{appendix:block-allocation}.

\subsection{Techniques for Practical Use}
List decoding is a well-established field in coding theory that decodes a list of messages that are within a certain hamming distance~\citep{elias1991error, guruswami2008explicit, guruswami2004list}. Inspired by this, we alter our decoding function to output candidate messages sorted by the level of confidence. In practice, list decoding is especially useful because provenance tracing via watermarking is far from finding an exact solution, but narrowing down the possible leakage points for a more detailed inspection that may be costly. For instance, when watermarking the timestamp of activity, it is useful to have a likely set of timestamps for which the practitioners to manually inspect, rather than a single candidate.  

Denoting the predicted message for position $i$ by $\hat{m}$, and the observed number of tokens in the colored list (strength of the watermark) by $w=\mathbf{W}_{i}[\hat{m}]$, the confidence of $\hat{m}$ should be higher if $w$ deviates from the expected mean under the null hypothesis that all colored lists are equally likely to be sampled. We define confidence at position $i$ as $c_i \propto {\text{Pr}(W^{\text{max}}_{i} \leq w | H_0)}$ where $W^\text{max}_i$ is the maximum cell value of $W_i\distas{H_0}\text{Multinomial}(T_i, [\gamma \cdots \gamma])$ where $T_i$ is the number of tokens assigned to position $i$. The distribution of $W^\text{max}_i$ is approximated using techniques from \citet{levin1981representation} (See Appendix \ref{appendix:max-multi-approx}).

Our algorithm can be parameterized by the confidence bound on each position:
\vspace{-1mm}
{\setlength{\leftmargini}{.5cm}   
\begin{itemize}
    \item Input: Best prediction $\hat{\mathbf{m}}$ found by majority voting via Alg. \ref{alg:extraction}, confidence bound $c_0$ 
    \vspace{-1mm}
    \item Output: $\hat{\mathbf{m}}_1,\cdots, \hat{\mathbf{m}}_{|\mathbb{L}|} \in \mathbb{L}$ whose predictions are altered on positions with confidence under $c_0$ 
    \end{itemize}
}
Empirically, we determine $c_0$ by constraining $|\mathbb{L}|$. Note that since $\hat{\textbf{m}}$ is always the most confident message, we comprise $\mathbb{L}$ with the next confident messages. To do this, we greedily alter the positions with the lowest confidence to the colorlist with the second largest number of tokens. Note that this list decoding technique is not unique to ours and can be applied to other methods as long as the decoding stage is computationally feasible.

This technique is not unique to ours and can be applied to other methods as long as the decoding stage is computationally feasible. Moreover, applying other techniques from coding theory such as error correction codes~\cite{wicker1999reed} to \ours~is straightforward as we independently encode the position of the message. More examples are in Appendix \ref{appendix:list-decoding}.

\section{Experiments}
    \label{sec:exp}
    \subsection{Experimental Settings}
For our main experiments, we use LLaMA-2-7B \citep{touvron2023llama} to generate sequences using  the newslike subset of the Colossal Common Crawl Cleaned corpus (C4) dataset~\citep{raffel2020exploring} following previous work~\citep{kirchenbauer2023watermark}. This simulates the scenario of generating fake news given a certain topic. For watermarking and text generation, we follow the configurations used in \citet{kirchenbauer2023reliability} unless otherwise denoted: bias $\delta=2.0$, greenlist ratio $\gamma=0.25$, which have shown a good trade-off between the detection performance and generation quality. Since $\gamma=0.25$, the number of colors $r$ is 4. We embed a random $b$-bit message onto $>$500 samples and report the mean metrics across samples.

When using the term `bit' or `bit-width', this denotes the initial message length; the effective message length is determined by $r$. When necessary, we also show the three standard error ranges. More details are in Appendix \ref{appendix:imp}.

\noindent\textbf{Metrics} To measure the performance of multi-bit watermarking, we use bit accuracy following previous works in the watermarking literature~\citep{zhu2018hidden, luo2020distortion, yang2022tracing, yoo2023robust} to measure how much of the embedded bits can be extracted without error. For zero-bit watermark performance (i.e. machine-text detection), we use area under the ROC curve (AUROC) and the true positive rate (TPR) at various false positive rate thresholds. Since all the baselines and ours are implemented on top of \texttt{Greenlist}, the impact on the text distribution is equivalent for all the methods when $\delta$ is equal. While \texttt{MSG-HASH} propose to use an auxiliary language model to aid the partitioning in the vocabulary, this is not feasible when the tokenizers of the main language model and the auxiliary one are different. Thus, we use the 'Vanilla Marking' proposed in their paper. We further discuss the validity of the metrics in Appendix \ref{appendix:bit-acc-as-metric}. To compute the performance of list decoding, we take the closest message out of the candidates.

\noindent\textbf{Threat Model} In the real world, a user may edit the generated text before publishing to refine the language model output or in an attempt to evade the watermark. We study two types of attacks studied in the past work~\citep{kirchenbauer2023reliability}: \textit{copy-paste} mixes the watermarked text and human text and \textit{paraphrasing} uses another language model to paraphrase the watermarked text. For the copy-paste attack, we randomly interleave the generated watermarked text into a non-watermarked text, mixing a $p$ percentage of non-watermarked texts while maintaining the total length. For paraphrasing, we use GPT-3.5-turbo (the prompt is shown in Table \ref{tab:robustness-gpt}). Both attacks do not maintain the start and end tokens of the watermarked text.

\begin{figure*}[t!]
    \begin{minipage}[t]{.75\textwidth}
        \centering
        \includegraphics[width=\textwidth]{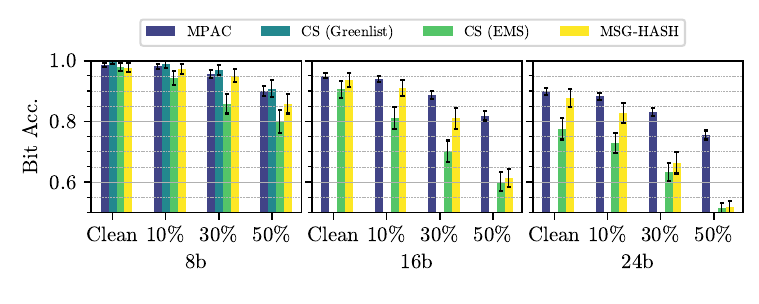}   
    \end{minipage}\hfill
    \begin{minipage}[t]{.25\textwidth}
        \includegraphics[width=\textwidth,left]{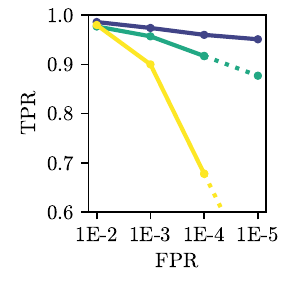}
    \end{minipage}
    \vspace{-10mm}
    \caption{\textit{Left}: Comparison with prior works without corruption (clean) and in the presence of copy-paste attack with $p$\%. On 24-bit, only 100 samples were watermarked for \cs~and \mh~due to lengthened encoding / decoding time. \textit{Right}: TPR for various FPR thresholds.}\label{fig:main-comparison}
    \vspace{-5mm}
\end{figure*}

\begin{figure}
\centering
\begin{minipage}[ht]{0.2\textwidth}
    \centering
    \includegraphics[width=\textwidth]{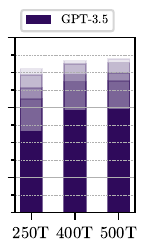}
\end{minipage}
\vspace{-4mm}
\caption{Corrupted bit accuracy for paraphrasing attack using GPT-3.5 embedding 8-bit messages at varying token lengths. We show multiple sizes of list ($|L|\in$\{2, 4, 8, 16\}) by color gradation as 8-bit has relatively small output space.}
\label{fig:gpt-robust}
\vspace{-4mm}
\end{figure}

\subsection{Results} \label{subsec:results}

For numerical results, see the tables in Appendix \ref{appendix:tables}.
    
\noindent\textbf{Comparison with Other Works.} We compare \ours~with \citet[\cs]{fernandez2023three} and \citet[\mh]{wang2023towards}. We do not compare with other steganography and post-processing works as they are extremely fragile in real-world corruption settings. Please refer to Sec.~\ref{sec:related} for details.
For \cs, the bit-width is bounded by $\log_2|\mathcal{V}|\approx$15 bits, since the cyclic-shift operation is only unique up to the size of the vocabulary. Due to this, we also experiment with extending \cs~to another zero-bit watermark method scheme called exponential minimum sampling \citep[EMS]{openai-watermark}, which does not have a theoretical upperbound. We call this \cs~(EMS).

The results in Fig. \ref{fig:main-comparison} show the clean and robust multi-bit accuracy in the presence of the copy-paste attack. At 8-bit, all methods achieve nearly 100\% accuracy and do fairly well even in the presence of corruption. At higher bit-width, \ours~outperforms others in both clean and robust accuracy. As corruption rate is increased, the other methods show dramatic degradation. In contrast, \ours~can withstand them due to position allocation, which independently encodes each position. In Fig. \ref{fig:main-comparison} Right, we compare the watermark detection performance at 8-bit. For \cs~ and \mh~, we use 10,000 negative samples and the TPRs@FPR=$\expnumber{1}{-5}$ are linearly interpolated due to the lengthened decoding time. The results demonstrate that \ours~outperforms them at low FPR thresholds. Notably, at FPR=$1e$-5, our true positive rate is .951. 

\begin{figure*}
    \begin{minipage}[h]{0.63\textwidth}
    \centering
    \includegraphics[width=\textwidth]{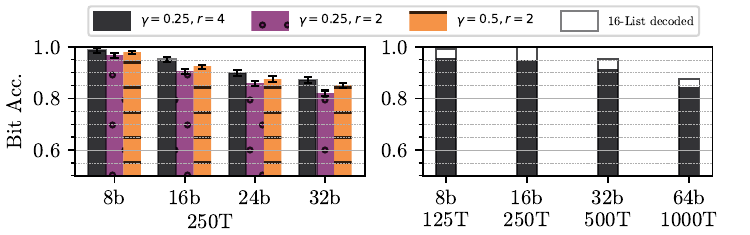}
    \subcaption{}
    \vspace{-5mm}
    \end{minipage}
    \begin{minipage}[h]{0.35\textwidth}
    \vspace{18pt}
    \centering
    \includegraphics[width=\textwidth]{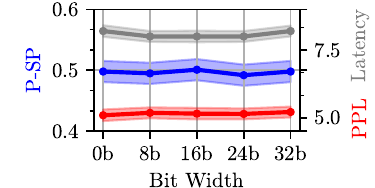}
    \subcaption{}
    \end{minipage}

\caption{(a) Clean bit accuracy with 3 standard errors for a fixed number of tokens (left) and fixed BPT (right). (b) Text quality (PPL, P-SP) and encoding latency across bit widths. 3 standard errors are shown.}\label{fig:ours-only}
\label{fig:robust}
\end{figure*}

Enlarging the message length comes at the cost of computation for prior works. Increasing the bit-width from 16-bit$\rightarrow$24-bit, lengthens the generation time of \cs~by roughly 3.6x (14 seconds $\rightarrow$ 50 seconds) per sample, while \ours~does not have increased latency (Fig. \ref{fig:ours-only}b). \mh~does not suffer from latency overhead during encoding, but the computation and memory overhead increase exponentially during decoding.  

\noindent\textbf{MPAC can maintain the watermark under various corruptions}.  
The full results of copy-paste attack in Appendix Fig. \ref{fig:robust-appendix}. Even at 32-bit, our watermark is not entirely destroyed as we encode each position of the watermark independently, which shows that it can benefit from error correction codes. We found paraphrasing to be much more challenging than the copy-paste attack when paraphrasing a short segment (T=$250$) and thus, experimented with only 8-bit messages and increasing the token lengths (Fig. \ref{fig:gpt-robust}). 
Nonetheless, the trends in robustness in the copy-paste attack remains the same with~\ours~~performing better than the baselines (Tab. \ref{tab:gpt-paraphrasing}). With T=$500$, the bit accuracy reaches nearly 80\% and with 16-list decoding, we are able to attain 90\% bit accuracy across all token lengths. More attacks using a different paraphraser model are considered in Appendix \ref{appendix:dipper}.

\begin{figure*}
    \begin{minipage}[t]{.32\textwidth}
    \centering
    \includegraphics[width=\textwidth]{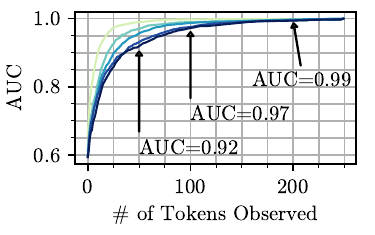}
    \subcaption{}\label{fig:detection}
    \end{minipage}
    \begin{minipage}[t]{.32\textwidth}
    \centering
    \includegraphics[width=\textwidth]{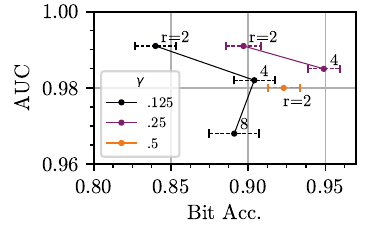}
    \subcaption{}\label{fig:gamma-radix}
    \end{minipage}
    \begin{minipage}[t]{.32\textwidth}
    \centering
    \includegraphics[width=\textwidth]{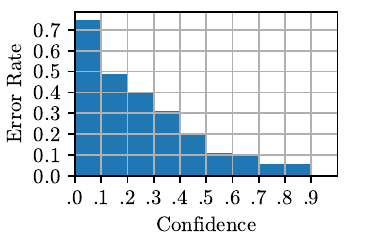}
    \subcaption{}\label{fig:confidence-vs-error}
    \end{minipage}
\vspace{-1mm}
\caption{(a) AUC@number of tokens observed for $b$=$\{0,8,16,24,32\}$. Darker colors denote larger bit-widths. (b) Zero-bit and multi-bit watermark performance for varying $\gamma$ and $r$ for 1000 samples at T=100,b=8. (c) Error rate as a function of confidence.}
\vspace{-3mm}
\end{figure*}

\noindent\textbf{Colorlisting improves multibit performance}. Next, we verify the effectiveness of `colorlisting', which takes advantage of the surplus vocabulary partitions. Fig. \ref{fig:ours-only}a demonstrates the gain in the load capacity by using $r$=$4$ colorlists as opposed to $r$=$2$ given a fixed $\gamma$. Besides the 8-bit case, which already achieves high accuracy, the performance of $\gamma=0.25$, $r$=$4$ is statistically significant at p=$\expnumber{1}{-2}$ than the second best variant. We further discuss the implications of varying $\gamma,r$ in Section \ref{sec:discussion}. 

Next, we increase the number of tokens (T) and bit width accordingly to demonstrate the feasibility of embedding longer messages. While the performance degrades as we increase the bit-width, the watermark does not entirely break, demonstrating the benefits of decomposing the message by positions. Moreover, the degradation can be partially compensated for by using list decoding. For 32-bit, the best possible message in the list achieves 95\% bit acc. by verifying only 16 out of $2^{32}$ possible messages. For 64-bit, the absolute performance gain is 3.0\% by generating merely 16 more candidate messages, which corresponds to roughly $1\mathrm{e}^{-20}$ of the total possible messages. Excluding the 8-bit case, whose AUC=$.988$, all the others have AUC $>.99$.

\noindent\textbf{Detection performance is affected by bit-width.} To get a clearer picture of the detection performance, we compute AUC vs. the number of tokens observed in Fig. \ref{fig:detection} following \citet{kirchenbauer2023reliability}. We see that the detection performance decreases as the message bit is increased. This phenomenon is similarly observed in other works as the increase in the number of ``hypotheses" required to check leads to an increase in the false positive rate~\citep{fernandez2023stable}. We further discuss the reasons behind this in the subsequent section. Note, however, that a watermarked text with 32-bit message reaches AUC over 0.99 once observing 200 tokens ($\approx$ 150 words). The TPR at FPR=$\expnumber{1}{-3}$ for b=$\{0,8,16,24,32\}$ are 0.98, 0.98, 0.95, 0.93, and 0.91, respectively (shown in Table \ref{tab:tpr}).

\noindent\textbf{Text quality is not affected by bit-width}. \ours~extends zero-bit watermarking by allocating tokens to message positions and partitioning vocabularies, which would otherwise be allocated to a single position and a single vocabulary partition. Consequently, given the same $\delta$ and $\gamma$, it only alters the text distribution to an extent that zero-bit watermarking does regardless of the bit-width. Indeed, our empirical results in Fig. \ref{fig:ours-only}b demonstrate that the text quality is statistically indistinguishable across bit-widths. We also show that the encoding latency, which directly experiences user experience, does not increase with bit-width. Three standard error ranges are shown.

\noindent\textbf{Across Model Scales, Datasets, Hash Schemes.} We further experiment with other pretrained models~\cite{jiang2023mistral, zhang2022opt} and their finetuned versions in Table \ref{tab:sft-rlhf}. The results demonstrate Mistral and OPT also achieve a similar performance, showing that our method is not limited to a specific pretrained model. We also find that the finetuned versions are also capable of watermarking, though the finetuned LLaMA model show a slight drop-off. The results for larger models (13B, 70B) and other datasets are in Appendix \ref{appendix:models}. 
To summarize, we found that text distributions with low entropy inherently have lower load capacity as observed similarly in prior works. However, our results consistently show that multi-bit watermarking is possible for open-form generation -- which resembles disinformation generation -- across model types and scales. We also present results for using another hash scheme with a longer context width in Appendix Table \ref{tab:fixedT} and \ref{tab:bpt}, which show a similar level of performance.

\begin{table}[t]
    \centering
    \begin{tabular}{cc}
        \toprule
        Model & Bit Acc. \\ \hline
        LLaMA-2-7b & .986 (.06) \\
        + Chat &  .922 (.13)   \\ \hline
        Mistral-7b & .987 (.06) \\
        + Chat & .977 (.08) \\ \hline
        OPT-1.3b &.982 (.07) \\
        \bottomrule
    \end{tabular}
    \caption{Performance on other pretrained models and their SFT and RLHF variants (\textit{Llama-2-7b-chat-hf} and \textit{Mistral-7B-Instruct-v0.1}). Results on $b$=8, $T$=250.}\label{tab:sft-rlhf}
\end{table}

\section{Discussions}\label{sec:discussion}
\vspace{-3mm}
\textbf{Load capacity and detection performance trade-off.} As noted above, embedding longer messages degrades the watermark detection performance due to overestimating the statistics of non-watermarked human texts (Fig. \ref{fig:z-at-t}). This is because computing the statistics involved finding the maximum cell value for each position. One natural solution is to use a better statistic that models the maximum cell value of a multinomial distribution. Empirically, we found that this performed on par or even slightly worse compared to the current approach, which may be due to the approximation error when using a small sample size. We give a more detailed discussion on this in Appendix \ref{appendix:detection-analysis}. 


\noindent\textbf{Radix and Colorlist proportion} 
How do radix and colorlist proportion $\gamma$ influence multi-bit watermark performance? For $\gamma$=.125, the benefits of enlarging $r$ to 8 are saturated and show no statistical significance to $r$=4. While larger $r$ allows more tokens to be assigned to each position by reducing the effective length of the message, it challenges the problem by increasing the number of possible answers (digits) per position. Additionally, we observed that increasing radix trade-offs zero-bit performance for multi-bit performance. The observations are illustrated in Fig. \ref{fig:gamma-radix}. 

\noindent\textbf{List Decoding Ablation}
In Fig. \ref{fig:confidence-vs-error}, we show a plot of bit error rate stratified by the confidence. While not properly calibrated (under-estimation), having higher confidence leads to lower error rate. We also highlight the effectiveness of this technique by comparing it with randomly outputting candidate messages from scratch in Table \ref{tab:mic}. We also observed that randomly altering a single position provides a good list as the best candidate message is already a good starting point. 

\section{Conclusion}
    \label{sec:conclusion}
    \vspace{-2mm}
Our findings demonstrate the effectiveness of embedding multi-bit message by allocating tokens to sub-units of the message. We show that this does not lead to quality degradation compared to its zero-bit counterpart nor latency overhead as bit-width is increased. This unveils a novel prospect of counteracting high-stake misuse of large language models via API. We also analyze the trade-off between multi-bit watermark and the detection rate.

\newpage

\section*{Limitations}
Watermarking is an prospective technology that can contribute to a safer use of large language models through promoting accountability. However, it is not yet a ready-made solution that can be deployed in the wild. One clear aspect that needs further study is quality compared to the non-watermarked generations. Recent empirical results demonstrate that watermarking generally leads to degradation in the quality~\citep{tu2023waterbench}. Several works attempt to tackle this by using techniques such as smaller language models to evenly partition the watermark~\citep{wang2023towards} or adaptive watermarks based on entropy threshold~\cite{lee2023wrote}. Since our proposed method MPAC can maintain the quality as its zero-bit counterpart, these improvements can be easily extended to our multi-bit framework as well. Another line of research propose different sampling strategies such as the exponential minimum sampling~\cite{openai-watermark} or non-distortionary functions~\citep{hu2023unbiased}. We believe the general framework of decomposing multi-bit message into units is easily generalizable to other watermarking scheme. An example is shown in Appendix \cref{appendix:block-allocation}.  

Another limitation of multi-bit watermark is its trade-off between the load capacity and detection rate (zero-bit watermark). While ours outperform the baselines in the zero-bit watermark setting (\cref{sec:exp}), overhauling this inherent limitation of multi-bit watermark remains another challenge in deploying multi-bit watermark over its zero-bit counterpart.

\section*{Ethics Statement}
Watermarking can mitigate malicious use cases by being able to trace back to the malicious user. This will enable holding accountability on adversaries for their malfeasance. However, ordinary users may find the idea discomforting as it may give the sense that the API provider can know what outputs are fed to the individual users. This is not the case unless the content is published to the public by the user, which -- in many cases -- is already done in an environment where the user can be identified (e.g. social media). All in all, the identification of machine-generated texts and tracing their provenance can enhance the accountability of API access of large language models without breaching individual users' privacy.

\section*{Acknowledgements}
This work was supported by IITP grant (2021-0-01343) and NRF grants (2021R1A2C3006659, 2022R1A5A7026673), all funded by MSIT of the Korean Government. 
    
\bibliography{custom}

\begin{thebibliography}{55}
\expandafter\ifx\csname natexlab\endcsname\relax\def\natexlab#1{#1}\fi

\bibitem[{Aaronson and Kirchner(2023)}]{openai-watermark}
Scott Aaronson and Hendrik Kirchner. 2023.
\newblock Watermarking gpt outputs.
\newblock \url{https://www.scottaaronson.com/talks/watermark.ppt}.
\newblock Accessed: 2023-09-14.

\bibitem[{Abdelnabi and Fritz(2021)}]{abdelnabi2021adversarial}
Sahar Abdelnabi and Mario Fritz. 2021.
\newblock Adversarial watermarking transformer: Towards tracing text provenance with data hiding.
\newblock In \emph{2021 IEEE Symposium on Security and Privacy (SP)}, pages 121--140. IEEE.

\bibitem[{Annie(2023)}]{amazon-reviews}
Palmer Annie. 2023.
\newblock \href {https://www.cnbc.com/2023/04/25/amazon-reviews-are-being-written-by-ai-chatbots.html} {People are using a.i. chatbots to write amazon reviews}.
\newblock \emph{CNBC}.

\bibitem[{Asikuzzaman and Pickering(2017)}]{asikuzzaman2017overview}
Md~Asikuzzaman and Mark~R Pickering. 2017.
\newblock An overview of digital video watermarking.
\newblock \emph{IEEE Transactions on Circuits and Systems for Video Technology}, 28(9):2131--2153.

\bibitem[{Atallah et~al.(2001)Atallah, Raskin, Crogan, Hempelmann, Kerschbaum, Mohamed, and Naik}]{atallah2001natural}
Mikhail~J Atallah, Victor Raskin, Michael Crogan, Christian Hempelmann, Florian Kerschbaum, Dina Mohamed, and Sanket Naik. 2001.
\newblock Natural language watermarking: Design, analysis, and a proof-of-concept implementation.
\newblock In \emph{International Workshop on Information Hiding}, pages 185--200. Springer.

\bibitem[{Badawy et~al.(2018)Badawy, Ferrara, and Lerman}]{Badway2018}
Adam Badawy, Emilio Ferrara, and Kristina Lerman. 2018.
\newblock \href {https://doi.org/10.1109/ASONAM.2018.8508646} {Analyzing the digital traces of political manipulation: The 2016 russian interference twitter campaign}.
\newblock In \emph{2018 IEEE/ACM International Conference on Advances in Social Networks Analysis and Mining (ASONAM)}, pages 258--265.

\bibitem[{Berlekamp(1964)}]{berlekamp1964block}
Elwyn~R Berlekamp. 1964.
\newblock \emph{Block coding with noiseless feedback}.
\newblock Ph.D. thesis, Massachusetts Institute of Technology.

\bibitem[{Cheddad et~al.(2010)Cheddad, Condell, Curran, and Mc~Kevitt}]{cheddad2010digital}
Abbas Cheddad, Joan Condell, Kevin Curran, and Paul Mc~Kevitt. 2010.
\newblock Digital image steganography: Survey and analysis of current methods.
\newblock \emph{Signal processing}, 90(3):727--752.

\bibitem[{Christ et~al.(2023)Christ, Gunn, and Zamir}]{christ2023undetectable}
Miranda Christ, Sam Gunn, and Or~Zamir. 2023.
\newblock Undetectable watermarks for language models.
\newblock \emph{arXiv preprint arXiv:2306.09194}.

\bibitem[{Cover(1999)}]{cover1999elements}
Thomas~M Cover. 1999.
\newblock \emph{Elements of information theory}.
\newblock John Wiley \& Sons.

\bibitem[{de~Witt et~al.(2023)de~Witt, Sokota, Kolter, Foerster, and Strohmeier}]{de2022perfectly}
Christian~Schroeder de~Witt, Samuel Sokota, J~Zico Kolter, Jakob~Nicolaus Foerster, and Martin Strohmeier. 2023.
\newblock Perfectly secure steganography using minimum entropy coupling.
\newblock In \emph{The Eleventh International Conference on Learning Representations}.

\bibitem[{Elias(1991)}]{elias1991error}
Peter Elias. 1991.
\newblock Error-correcting codes for list decoding.
\newblock \emph{IEEE Transactions on Information Theory}, 37(1):5--12.

\bibitem[{Fang et~al.(2017)Fang, Jaggi, and Argyraki}]{fang2017generating}
Tina Fang, Martin Jaggi, and Katerina Argyraki. 2017.
\newblock Generating steganographic text with lstms.
\newblock In \emph{Proceedings of ACL 2017, Student Research Workshop}, pages 100--106.

\bibitem[{Fernandez et~al.(2023{\natexlab{a}})Fernandez, Chaffin, Tit, Chappelier, and Furon}]{fernandez2023three}
Pierre Fernandez, Antoine Chaffin, Karim Tit, Vivien Chappelier, and Teddy Furon. 2023{\natexlab{a}}.
\newblock Three bricks to consolidate watermarks for large language models.
\newblock \emph{arXiv preprint arXiv:2308.00113}.

\bibitem[{Fernandez et~al.(2023{\natexlab{b}})Fernandez, Couairon, J{\'e}gou, Douze, and Furon}]{fernandez2023stable}
Pierre Fernandez, Guillaume Couairon, Herv{\'e} J{\'e}gou, Matthijs Douze, and Teddy Furon. 2023{\natexlab{b}}.
\newblock The stable signature: Rooting watermarks in latent diffusion models.
\newblock \emph{arXiv preprint arXiv:2303.15435}.

\bibitem[{Gage(1994)}]{gage1994new}
Philip Gage. 1994.
\newblock A new algorithm for data compression.
\newblock \emph{C Users Journal}, 12(2):23--38.

\bibitem[{Group(2018)}]{posix}
The~Open Group. 2018.
\newblock The open group base specifications issue 7, 2018 edition ieee std 1003.1™-2017 (revision of ieee std 1003.1-2008) copyright © 2001 2018 ieee and the open group.
\newblock \url{https://pubs.opengroup.org/onlinepubs/9699919799/}.
\newblock Accessed: 2023-09-14.

\bibitem[{Gupta et~al.(2023)Gupta, Guruswami, and Zhang}]{gupta2023binary}
Meghal Gupta, Venkatesan Guruswami, and Rachel~Yun Zhang. 2023.
\newblock Binary error-correcting codes with minimal noiseless feedback.
\newblock In \emph{Proceedings of the 55th Annual ACM Symposium on Theory of Computing}, pages 1475--1487.

\bibitem[{Guruswami(2004)}]{guruswami2004list}
Venkatesan Guruswami. 2004.
\newblock \emph{List decoding of error-correcting codes: winning thesis of the 2002 ACM doctoral dissertation competition}, volume 3282.
\newblock Springer Science \& Business Media.

\bibitem[{Guruswami and Rudra(2008)}]{guruswami2008explicit}
Venkatesan Guruswami and Atri Rudra. 2008.
\newblock Explicit codes achieving list decoding capacity: Error-correction with optimal redundancy.
\newblock \emph{IEEE Transactions on information theory}, 54(1):135--150.

\bibitem[{He et~al.(2022)He, Xu, Lyu, Wu, and Wang}]{he2022protecting}
Xuanli He, Qiongkai Xu, Lingjuan Lyu, Fangzhao Wu, and Chenguang Wang. 2022.
\newblock Protecting intellectual property of language generation apis with lexical watermark.
\newblock In \emph{Proceedings of the AAAI Conference on Artificial Intelligence}, volume~36, pages 10758--10766.

\bibitem[{Hu(2023)}]{chatgpt-users}
Krystal Hu. 2023.
\newblock \href {https://www.reuters.com/technology/chatgpt-sets-record-fastest-growing-user-base-analyst-note-2023-02-01/} {Chatgpt sets record for fastest-growing user base - analyst note}.
\newblock \emph{Reuters}.

\bibitem[{Hu et~al.(2023)Hu, Chen, Wu, Wu, Zhang, and Huang}]{hu2023unbiased}
Zhengmian Hu, Lichang Chen, Xidong Wu, Yihan Wu, Hongyang Zhang, and Heng Huang. 2023.
\newblock Unbiased watermark for large language models.
\newblock \emph{arXiv preprint arXiv:2310.10669}.

\bibitem[{Hua et~al.(2016)Hua, Huang, Shi, Goh, and Thing}]{hua2016twenty}
Guang Hua, Jiwu Huang, Yun~Q Shi, Jonathan Goh, and Vrizlynn~LL Thing. 2016.
\newblock Twenty years of digital audio watermarking—a comprehensive review.
\newblock \emph{Signal processing}, 128:222--242.

\bibitem[{Jiang et~al.(2023)Jiang, Sablayrolles, Mensch, Bamford, Chaplot, Casas, Bressand, Lengyel, Lample, Saulnier et~al.}]{jiang2023mistral}
Albert~Q Jiang, Alexandre Sablayrolles, Arthur Mensch, Chris Bamford, Devendra~Singh Chaplot, Diego de~las Casas, Florian Bressand, Gianna Lengyel, Guillaume Lample, Lucile Saulnier, et~al. 2023.
\newblock Mistral 7b.
\newblock \emph{arXiv preprint arXiv:2310.06825}.

\bibitem[{Kirchenbauer et~al.(2023{\natexlab{a}})Kirchenbauer, Geiping, Wen, Katz, Miers, and Goldstein}]{kirchenbauer2023watermark}
John Kirchenbauer, Jonas Geiping, Yuxin Wen, Jonathan Katz, Ian Miers, and Tom Goldstein. 2023{\natexlab{a}}.
\newblock A watermark for large language models.
\newblock \emph{arXiv preprint arXiv:2301.10226}.

\bibitem[{Kirchenbauer et~al.(2023{\natexlab{b}})Kirchenbauer, Geiping, Wen, Shu, Saifullah, Kong, Fernando, Saha, Goldblum, and Goldstein}]{kirchenbauer2023reliability}
John Kirchenbauer, Jonas Geiping, Yuxin Wen, Manli Shu, Khalid Saifullah, Kezhi Kong, Kasun Fernando, Aniruddha Saha, Micah Goldblum, and Tom Goldstein. 2023{\natexlab{b}}.
\newblock On the reliability of watermarks for large language models.
\newblock \emph{arXiv preprint arXiv:2306.04634}.

\bibitem[{Krishna et~al.(2023)Krishna, Song, Karpinska, Wieting, and Iyyer}]{krishna2023paraphrasing}
Kalpesh Krishna, Yixiao Song, Marzena Karpinska, John Wieting, and Mohit Iyyer. 2023.
\newblock Paraphrasing evades detectors of ai-generated text, but retrieval is an effective defense.
\newblock \emph{arXiv preprint arXiv:2303.13408}.

\bibitem[{Kuditipudi et~al.(2023)Kuditipudi, Thickstun, Hashimoto, and Liang}]{kuditipudi2023robust}
Rohith Kuditipudi, John Thickstun, Tatsunori Hashimoto, and Percy Liang. 2023.
\newblock Robust distortion-free watermarks for language models.
\newblock \emph{arXiv preprint arXiv:2307.15593}.

\bibitem[{Lee et~al.(2023)Lee, Hong, Ahn, Hong, Lee, Yun, Shin, and Kim}]{lee2023wrote}
Taehyun Lee, Seokhee Hong, Jaewoo Ahn, Ilgee Hong, Hwaran Lee, Sangdoo Yun, Jamin Shin, and Gunhee Kim. 2023.
\newblock Who wrote this code? watermarking for code generation.
\newblock \emph{arXiv preprint arXiv:2305.15060}.

\bibitem[{Levin(1981)}]{levin1981representation}
Bruce Levin. 1981.
\newblock A representation for multinomial cumulative distribution functions.
\newblock \emph{The Annals of Statistics}, pages 1123--1126.

\bibitem[{Luo et~al.(2020)Luo, Zhan, Chang, Yang, and Milanfar}]{luo2020distortion}
Xiyang Luo, Ruohan Zhan, Huiwen Chang, Feng Yang, and Peyman Milanfar. 2020.
\newblock Distortion agnostic deep watermarking.
\newblock In \emph{Proceedings of the IEEE/CVF Conference on Computer Vision and Pattern Recognition}, pages 13548--13557.

\bibitem[{Merity et~al.(2016)Merity, Xiong, Bradbury, and Socher}]{merity2016pointer}
Stephen Merity, Caiming Xiong, James Bradbury, and Richard Socher. 2016.
\newblock \href {http://arxiv.org/abs/1609.07843} {Pointer sentinel mixture models}.

\bibitem[{Mitchell et~al.(2023)Mitchell, Lee, Khazatsky, Manning, and Finn}]{mitchell2023detectgpt}
Eric Mitchell, Yoonho Lee, Alexander Khazatsky, Christopher~D Manning, and Chelsea Finn. 2023.
\newblock Detectgpt: Zero-shot machine-generated text detection using probability curvature.
\newblock \emph{arXiv preprint arXiv:2301.11305}.

\bibitem[{Munyer and Zhong(2023)}]{munyer2023deeptextmark}
Travis Munyer and Xin Zhong. 2023.
\newblock Deeptextmark: Deep learning based text watermarking for detection of large language model generated text.
\newblock \emph{arXiv preprint arXiv:2305.05773}.

\bibitem[{Pierri et~al.(2023)Pierri, Luceri, Jindal, and Ferrara}]{pierri2023propaganda}
Francesco Pierri, Luca Luceri, Nikhil Jindal, and Emilio Ferrara. 2023.
\newblock Propaganda and misinformation on facebook and twitter during the russian invasion of ukraine.
\newblock In \emph{Proceedings of the 15th ACM Web Science Conference 2023}, pages 65--74.

\bibitem[{Potdar et~al.(2005)Potdar, Han, and Chang}]{potdar2005survey}
Vidyasagar~M Potdar, Song Han, and Elizabeth Chang. 2005.
\newblock A survey of digital image watermarking techniques.
\newblock In \emph{INDIN'05. 2005 3rd IEEE International Conference on Industrial Informatics, 2005.}, pages 709--716. IEEE.

\bibitem[{Raffel et~al.(2020)Raffel, Shazeer, Roberts, Lee, Narang, Matena, Zhou, Li, and Liu}]{raffel2020exploring}
Colin Raffel, Noam Shazeer, Adam Roberts, Katherine Lee, Sharan Narang, Michael Matena, Yanqi Zhou, Wei Li, and Peter~J Liu. 2020.
\newblock Exploring the limits of transfer learning with a unified text-to-text transformer.
\newblock \emph{The Journal of Machine Learning Research}, 21(1):5485--5551.

\bibitem[{Schuhmann(2022)}]{essay}
Christoph Schuhmann. 2022.
\newblock Huggingface datasets: Christophschuhmann/essays-with-instructions.
\newblock \url{https://huggingface.co/datasets/ChristophSchuhmann/essays-with-instructions}.
\newblock Accessed: 2023-09-14.

\bibitem[{Topkara et~al.(2006{\natexlab{a}})Topkara, Riccardi, Hakkani-T{\"u}r, and Atallah}]{topkara2006natural}
Mercan Topkara, Giuseppe Riccardi, Dilek Hakkani-T{\"u}r, and Mikhail~J Atallah. 2006{\natexlab{a}}.
\newblock Natural language watermarking: Challenges in building a practical system.
\newblock In \emph{Security, Steganography, and Watermarking of Multimedia Contents VIII}, volume 6072, pages 106--117. SPIE.

\bibitem[{Topkara et~al.(2005)Topkara, Taskiran, and Delp~III}]{topkara2005natural}
Mercan Topkara, Cuneyt~M Taskiran, and Edward~J Delp~III. 2005.
\newblock Natural language watermarking.
\newblock In \emph{Security, Steganography, and Watermarking of Multimedia Contents VII}, volume 5681, pages 441--452. SPIE.

\bibitem[{Topkara et~al.(2006{\natexlab{b}})Topkara, Topkara, and Atallah}]{topkara2006hiding}
Umut Topkara, Mercan Topkara, and Mikhail~J Atallah. 2006{\natexlab{b}}.
\newblock The hiding virtues of ambiguity: quantifiably resilient watermarking of natural language text through synonym substitutions.
\newblock In \emph{Proceedings of the 8th workshop on Multimedia and security}, pages 164--174.

\bibitem[{Touvron et~al.(2023)Touvron, Martin, Stone, Albert, Almahairi, Babaei, Bashlykov, Batra, Bhargava, Bhosale et~al.}]{touvron2023llama}
Hugo Touvron, Louis Martin, Kevin Stone, Peter Albert, Amjad Almahairi, Yasmine Babaei, Nikolay Bashlykov, Soumya Batra, Prajjwal Bhargava, Shruti Bhosale, et~al. 2023.
\newblock Llama 2: Open foundation and fine-tuned chat models.
\newblock \emph{arXiv preprint arXiv:2307.09288}.

\bibitem[{Tu et~al.(2023)Tu, Sun, Bai, Yu, Hou, and Li}]{tu2023waterbench}
Shangqing Tu, Yuliang Sun, Yushi Bai, Jifan Yu, Lei Hou, and Juanzi Li. 2023.
\newblock Waterbench: Towards holistic evaluation of watermarks for large language models.
\newblock \emph{arXiv preprint arXiv:2311.07138}.

\bibitem[{Valenzuela et~al.(2022)Valenzuela, Halpern, and Araneda}]{valenzuela2022downward}
Sebasti{\'a}n Valenzuela, Daniel Halpern, and Felipe Araneda. 2022.
\newblock A downward spiral? a panel study of misinformation and media trust in chile.
\newblock \emph{The International Journal of Press/Politics}, 27(2):353--373.

\bibitem[{Wang et~al.(2023{\natexlab{a}})Wang, Yang, Chen, Zhou, Lin, Meng, Zhou, and Sun}]{wang2023towards}
Lean Wang, Wenkai Yang, Deli Chen, Hao Zhou, Yankai Lin, Fandong Meng, Jie Zhou, and Xu~Sun. 2023{\natexlab{a}}.
\newblock Towards codable text watermarking for large language models.
\newblock \emph{arXiv preprint arXiv:2307.15992}.

\bibitem[{Wang et~al.(2023{\natexlab{b}})Wang, Mansurov, Ivanov, Su, Shelmanov, Tsvigun, Whitehouse, Afzal, Mahmoud, Aji et~al.}]{wang2023m4}
Yuxia Wang, Jonibek Mansurov, Petar Ivanov, Jinyan Su, Artem Shelmanov, Akim Tsvigun, Chenxi Whitehouse, Osama~Mohammed Afzal, Tarek Mahmoud, Alham~Fikri Aji, et~al. 2023{\natexlab{b}}.
\newblock M4: Multi-generator, multi-domain, and multi-lingual black-box machine-generated text detection.
\newblock \emph{arXiv preprint arXiv:2305.14902}.

\bibitem[{Wicker and Bhargava(1999)}]{wicker1999reed}
Stephen~B Wicker and Vijay~K Bhargava. 1999.
\newblock \emph{Reed-Solomon codes and their applications}.
\newblock John Wiley \& Sons.

\bibitem[{Wieting et~al.(2022)Wieting, Gimpel, Neubig, and Berg-Kirkpatrick}]{wieting2022paraphrastic}
John Wieting, Kevin Gimpel, Graham Neubig, and Taylor Berg-Kirkpatrick. 2022.
\newblock Paraphrastic representations at scale.
\newblock In \emph{Proceedings of the The 2022 Conference on Empirical Methods in Natural Language Processing: System Demonstrations}, pages 379--388.

\bibitem[{Yang et~al.(2022)Yang, Zhang, Chen, Zhang, Ma, Wang, and Yu}]{yang2022tracing}
Xi~Yang, Jie Zhang, Kejiang Chen, Weiming Zhang, Zehua Ma, Feng Wang, and Nenghai Yu. 2022.
\newblock Tracing text provenance via context-aware lexical substitution.
\newblock In \emph{Proceedings of the AAAI Conference on Artificial Intelligence}, volume~36, pages 11613--11621.

\bibitem[{Yoo et~al.(2023)Yoo, Ahn, Jang, and Kwak}]{yoo2023robust}
KiYoon Yoo, Wonhyuk Ahn, Jiho Jang, and Nojun Kwak. 2023.
\newblock Robust multi-bit natural language watermarking through invariant features.
\newblock In \emph{Proceedings of the 61st Annual Meeting of the Association for Computational Linguistics (Volume 1: Long Papers)}, pages 2092--2115.

\bibitem[{Zhang et~al.(2022)Zhang, Roller, Goyal, Artetxe, Chen, Chen, Dewan, Diab, Li, Lin et~al.}]{zhang2022opt}
Susan Zhang, Stephen Roller, Naman Goyal, Mikel Artetxe, Moya Chen, Shuohui Chen, Christopher Dewan, Mona Diab, Xian Li, Xi~Victoria Lin, et~al. 2022.
\newblock Opt: Open pre-trained transformer language models.
\newblock \emph{arXiv preprint arXiv:2205.01068}.

\bibitem[{Zhao et~al.(2023)Zhao, Pang, Du, Yang, Cheung, and Lin}]{zhao2023recipe}
Yunqing Zhao, Tianyu Pang, Chao Du, Xiao Yang, Ngai-Man Cheung, and Min Lin. 2023.
\newblock A recipe for watermarking diffusion models.
\newblock \emph{arXiv preprint arXiv:2303.10137}.

\bibitem[{Zhu et~al.(2018)Zhu, Kaplan, Johnson, and Fei-Fei}]{zhu2018hidden}
Jiren Zhu, Russell Kaplan, Justin Johnson, and Li~Fei-Fei. 2018.
\newblock Hidden: Hiding data with deep networks.
\newblock In \emph{Proceedings of the European conference on computer vision (ECCV)}, pages 657--672.

\bibitem[{Ziegler et~al.(2019)Ziegler, Deng, and Rush}]{ziegler2019neural}
Zachary Ziegler, Yuntian Deng, and Alexander~M Rush. 2019.
\newblock Neural linguistic steganography.
\newblock In \emph{Proceedings of the 2019 Conference on Empirical Methods in Natural Language Processing and the 9th International Joint Conference on Natural Language Processing (EMNLP-IJCNLP)}, pages 1210--1215.

\end{thebibliography}

\newpage
\appendix
\section{Appendix}
    \label{sec:appendix}
    \paragraph{Table of Contents}\label{para:toc}
\begin{enumerate}
    \item \nameref{appendix:algo}
    \item \nameref{appendix:detection-analysis}
    \item \nameref{appendix:list-decoding}
    \item \nameref{appendix:block-allocation} 
    \item \nameref{appendix:imp}
    \item \nameref{appendix:bit-acc-as-metric}
    \item \nameref{appendix:hash}
    \item \nameref{appendix:dipper}
    \item \nameref{appendix:models}
    \item \nameref{appendix:tables}
    \item \nameref{appendix:generation samples}
\end{enumerate}
\clearpage

\subsection{Decoding Algorithm}\label{appendix:algo}
\SetKwInput{KwInput}{Input}              
\SetKwInput{KwOutput}{Output}  
\maketitle

\begin{algorithm}
    \DontPrintSemicolon
    \KwInput{Watermarked text $X_{1:T}$, hash context width $h$, effective message length $\tilde b$}
    \KwOutput{Predicted message $\hat{\textbf{m}}$, number of colorlisted tokens $w$}
    \tcc{Initialize counter}
    {$\textbf{W}_{p}[m] = 0 \text{   } \forall p,m$}\; 
    \tcc{Count tokens in colored lists}
    \For{$t \texttt{ in } [h+1, T]$}
    {
        $s=f(X_{t-h:t-1})$\;
        $p=\texttt{sample}([\tilde b])$ \;
        \For{$m \texttt{ in }[r]$}
        {
            Permute $\mathcal{V}_t$ using $s$ as seed\;
            \If{$X_t \in \mathcal{G}_t^{m}$}
                {
                $\textbf{W}_{p}[m]$ += 1 \;
                }
    
        }
        
    }
    \tcc{Predict message}
    $\hat{\textbf{m}}_r = \text{`` "}$\; 
    $w=0$\;
    \For{$p \texttt{ in } [\tilde b]$}
    {
        $w$ += $\texttt{max}(\textbf{W}_{p}[m])$\;
        $\hat{m}=\texttt{argmax}_m(\mathbf{W}_p[m])$\;
        $\hat{\textbf{m}}_r$ += $\texttt{str}(\hat{m}$) \; 
    }
    Get bit message $\hat{\textbf{m}}$ by converting $\hat{\textbf{m}}_r$ \;
    \Return $\hat{\textbf{m}}$, $w$
\caption{Message Decoding}
\label{alg:extraction-long}
\end{algorithm}

\begin{figure}
    \centering
    \includegraphics{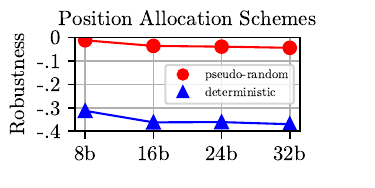}
    \caption{Performance difference between watermark extraction with and without corruption. "Deterministic" denotes sequentially encoding each position of the message as done in \citet{ziegler2019neural} in the \texttt{Greenlist} framework.
    Mixing 20\% of non-watermarked text makes the bit accuracy of sequential encoding scheme nearly random.}
    \label{fig:fragile-coding}
\end{figure}

\subsection{Analysis on Watermark Detection}\label{appendix:detection-analysis}
\subsubsection{Watermark Detection}
The presence of a watermark is determined by counting the number of tokens in the greenlist. For a human-generated text that has no knowledge of the greenlist rule, a token will be from the greenlist with the probability $\gamma \leq 0.5$, the proportion of the greenlist size compared to the entire vocabulary. Without the knowledge of the greenlist (null hypothesis), the number of tokens in the greenlist ($g$) follows a binomial distribution. \citep{kirchenbauer2023watermark} used the normal approximation to the binomial distribution to compute the $z$-statistics for a text with $T$ tokens: $z = \frac{g - \gamma T}{\sqrt{\gamma(1-\gamma)T}}$.

\begin{figure*}[t!]
    \centering
    \includegraphics{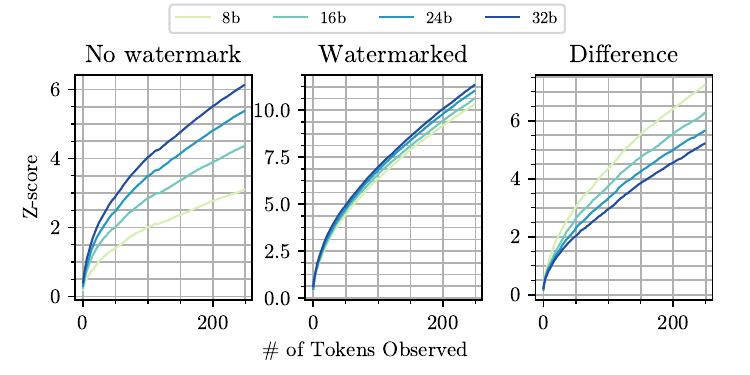}
    \caption{The detection scores of non-watermarked texts, watermarked texts and their difference as a function of number of tokens observed. We see that the difference in the scores decreases as bit-width increases, leading to reduced seperability.}
    \label{fig:z-at-t}
\end{figure*}

Next, we further analyze how bit-width of the message and radix affect detection performance. Our analysis stems from the observation that as we increase the bit-width the detection score for the non-watermarked text increases more rapidly than that of the watermarked text. Consequently, the difference in the two scores \textit{decreases} as larger bit-width is used, leading to reduced seperability. The results are in Fig. \ref{fig:z-at-t}. Notice that the difference between the scores of watermarked and non-watermarked texts decreases for larger bit-width.

To grasp a hint of what is going on, we do away with the language model and other complexities by modeling this only through statistical distributions. To recap, our detection statistic (Eq. \ref{eq:zero-bit-detection}) was computed by aggregating the number of tokens in each position of the message. Letting $C_i$ as the number of tokens in the colorlist for the position $i$, we can write the aggregated form as 
\begin{equation}
    C = C_0 + \cdots + C_{p-1} \distas{H_0} \text{Binomial}(T, \gamma)
\end{equation}

\begin{figure}[th!]
    \centering
    \includegraphics{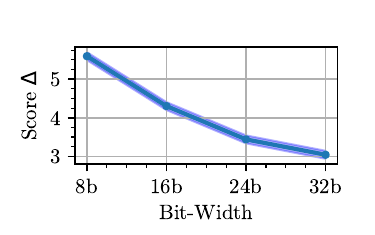}
    \includegraphics{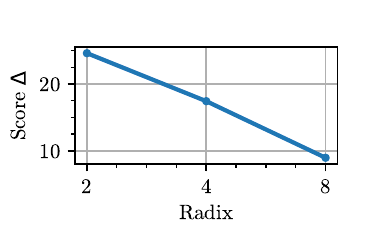}
    \caption{Simulation of the difference between (unormalized) scores for watermarked and non-watermarked multinomial distributions. Higher score signify higher seperability, hence higher detection performance. We use $\epsilon$=0.1. For right, we use $\gamma$=.125 to allow more radix.}\label{fig:simulation}
\end{figure}

However, note that during decoding the ground truth message is unknown and thus, is predicted by taking the colorlist that has the max number of tokens. This is problematic when decoding for non-watermarked text as it biases the statistic to be higher when bit-width is increased. We let $W_i=[w_0, \dots, w_{r-1}]$ be the number of tokens in $r$ colorlists (strength of watermark) for position $i$. For a non-watermarked text, we can assume that this is a random variable with equal probability for each colorlist 
\begin{equation}
    W_i\distas{} \text{Multinomial}(n_i, [\gamma \cdots \gamma])
\end{equation}
where $n_i$ is the number of tokens allocated to position $i$. Our decoding method takes the maximum cell value of this, which makes itself a random variable:
\begin{equation}\label{eq:radix}
    W^\text{max}_i = \mathrm{max}(W_i) =  \mathrm{max}([w_0,\dots,w_{r-1}]).
\end{equation}
Our final statistic used for our detection score is the sum of this variable over the entire positions: 
\begin{equation}\label{eq:pos}
    W^\text{max} = \sum_i^{p} W^\text{max}_i
\end{equation}
We see that our statistic is dependent upon the number of candidates when selecting the maximum cell (i.e. radix) through Eq. \ref{eq:radix} and the number of positions (i.e. bit-width) through Eq. \ref{eq:pos}.

To verify the effect of bit-width and radix on the detection score, we compare the difference in the statistics for a uniform multinomial distribution, which signify non-watermarked text, and a multinomial distribution with a slightly modified probability $[\gamma + \epsilon, \gamma, \dots ,\gamma]$ to signify the added bias term for the watermarked distribution. We sample 1000 samples of $W^\text{max}$ and compute the difference in the detection scores for the two distributions. The results in Fig. \ref{fig:simulation} corroborate that an increase in bit-width / radix decreases the separability of the detection scores.

In an attempt to overhaul this, we tried computing the likelihood of $W_i^\text{rm}$ before aggregating them using an approximation of \cite{levin1981representation} (More details in the next section). However, this only led to on par or slightly worse performance. This may be because $n_i$ is small for cases when $T$ is small compared to the length of the message. 
Other than this, some of the approaches we attempted were:
\begin{itemize}
    \item Computing test statistic per position or weighting the statistic of each position with $n_i$ before aggregating.
    \item Computing the p-value of the binomial random variables rather than using the normal approximation, i.e. regularized incomplete beta function.
    \item Computing the p-value under the null hypothesis that the distribution of the colorlists follows a uniform distribution, i.e. Chi-square Goodness of Fit test    
\end{itemize}
All the approaches either led to on-par or slightly worse results.

\subsubsection{Approximating Max Multinomial Cell Distribution}\label{appendix:max-multi-approx}
We used the approximation of \cite{levin1981representation} for modeling the distribution of the maximum cell frequency. For completeness, we present the steps used for the approximation adapted to our case. For a multinomial distribution with sample size $N$ and probability vectors $[p_0, \dots, p_{r-1}]$, Let $a$ be the maximum cell value, then the cumulative distribution function of having a maximum value of $a$ can be approximated for any real number $s > 0$
\begin{equation}
    P(a) = \frac{N!}{s^{N}e^{-s}}\{\prod_i^{r-1} P(X_i \leq a)\} P(W=N)
\end{equation}
where $X_i\distas{}$Poisson($sp_i$) and $W=\sum_i^{r-1}=Y_i\distas{}$Truncated Poisson($sp_i$) with range $0,1,\dots,a$. Following Example 1 of \cite{levin1981representation}, we set $s=N$ and use Stirling's approximation for $N!$. We also approximate $W$ using the normal approximation to the Poisson distribution.

\subsection{List Decoding and Other Techniques}\label{appendix:list-decoding} 
The decomposition of the message into each bit position bounds the computation during decoding to the number of tokens. This allows \ours~to output a \textit{list} of most likely messages without exhaustively considering all the possible messages. We alter our decoding function to output candidate messages sorted by the level of confidence. Denoting the predicted message for position $i$ by $\hat{m}$, and the observed number of tokens in the colored list (strength of the watermark) by $w=\mathbf{W}_{i}[\hat{m}]$, the confidence of $\hat{m}$ should be higher if $w$ deviates from the expected mean under the null hypothesis that all colored lists are equally likely to be sampled. We define confidence at position $i$ as $c_i \propto {\text{Pr}(W^{\text{max}}_{i} \leq w | H_0)}$ where $W^\text{max}_i$ is the maximum cell value of $W_i\distas{H_0}\text{Multinomial}(T_i, [\gamma \cdots \gamma])$ where $T_i$ is the number of tokens assigned to position $i$. The distribution of $W^\text{max}_i$ is approximated using techniques from \citet{levin1981representation} (See Appendix \ref{appendix:max-multi-approx}).

Our algorithm can be parameterized by the confidence bound on each position:
\vspace{-1mm}
{\setlength{\leftmargini}{.5cm}   
\begin{itemize}
    \item Input: Best prediction $\hat{\mathbf{m}}$ found by majority voting via Alg. \ref{alg:extraction}, confidence bound $c_0$ 
    \vspace{-1mm}
    \item Output: $\hat{\mathbf{m}}_1,\cdots, \hat{\mathbf{m}}_{|\mathbb{L}|} \in \mathbb{L}$ whose predictions are altered on positions with confidence under $c_0$ 
    \end{itemize}
}
Empirically, we determine $c_0$ by constraining $|\mathbb{L}|$. Note that since $\hat{\textbf{m}}$ is always the most confident message, we comprise $\mathbb{L}$ with the next confident messages. To do this, we greedily alter the positions with the lowest confidence to the colorlist with the second largest number of tokens. Note that this list decoding technique is not unique to ours and can be applied to other methods as long as the decoding stage is computationally feasible.

\begin{table}
    \centering   
    \begin{tabular}{c|ccc}
    \toprule

    \multicolumn{4}{c}{Bit Accuracy} \\ 
        $\delta$    & 0.5  &  1  &  2 \\
    \toprule
    No feedback & .626  &  .766  & .948  \\  
    $\tilde \delta=\delta$ + 1 & .769 & .860 & .960  \\
    \bottomrule
\end{tabular}
\caption{Results for using feedback for adapting bias on T=100,b=8}
\label{tab:adaptive-bias}
\end{table}

\begin{table}
\centering
\scriptsize
\begin{tabular}{c|cccc}
    \toprule
    \multicolumn{5}{c}{Accuracy Gained} \\ 
    &  8b &  16b  &  24b & 32b \\
    \toprule
    $c_i$-sorted list & 1.1\%  &  3.7\%  & 6.0\% & 5.6\%  \\
    
    Random list & 0.6\% & 0.4\% & 0.5\% &  0.3\% \\
    \bottomrule
\end{tabular}
\begin{tabular}{c|ccccc}
    \multicolumn{6}{c}{Latency (seconds$\slash$250 tokens )} \\
    & 0b & 8b &  16b  &  24b & 32b \\
    \toprule 
    Encoding (7.9) & 8.19 & 7.98 & 8.01 & 7.96 & 8.24  \\
    Decoding (.09) & .08 & .09 & .09 & .09 & .10 \\
    \bottomrule
\end{tabular}
\caption{Comparison of absolute improvement in bit accuracy when using confidence-based list decoding and random list.}\label{tab:mic}
\end{table}

\subsubsection{Results}
We show absolute accuracy gained using confidence-based list decoding ($|L|$=16) compared with random decoding. We further compare the encoding and decoding latency for sequences with $\sim 250$ tokens using a single Nvidia A100 when using an additive left hash scheme with context width 1. The results are in Table \ref{tab:mic}. The latency \textit{does not} proportionally increase with message bit length, making it scalable to long messages. When using an efficient hashing scheme watermarking has a negligible increase in both encoding and decoding compared to vanilla generation, which requires 7.9 seconds and 0.09 seconds, respectively. 

\subsubsection{Error Correction Codes}
We first use the basic notions from coding theory adapted from \citet{cover1999elements} to formulate our problem:  
\begin{itemize}
    \item Encoding function is a function $E: \mathcal{M} \rightarrow \mathcal{X}$  that maps the original message into longer, usually redundant string where $\mathcal{M} \subseteq [r]^b, \mathcal{X} \subseteq \Sigma^T$. The rate of $E$ is given by $\frac{b}{T} \log_2 r$ bits/symbol.
    \item  $p(y|x)$ is a noisy channel that models the transmission of the encoded message. 
    \item A channel's capacity is the upper bound of the rate of an encoding function in order for a reliable transmission.
   \item Decoding function is a function $D: \mathcal{Y} \rightarrow \mathcal{M}$ that recovers the original message from $y$.
\end{itemize}
We first simplify our setting to embedding a single-digit message ($b=1$), which does not lead to a loss of generality as \texttt{MPAC} encodes each position independently. Each token of a language model is a signal for embedding the message ($m$) by repetitively sampling from the $m^{\text{th}}$ colorlist. Therefore, our \textbf{encoding} function is a repetition code that maps a redundant message content $T$ (number of tokens) times. Our \textbf{channel} is the generation process of the language model, which stochastically transmits the encoded message by sampling from the vocabulary distribution that has been modified to favor the selected colorlist. The success probability of each transmission depends on the magnitude of the bias $\delta$, the entropy of the vocabulary distribution, and, more holistically, the text distribution. The \textbf{decoding} function selects the argmax of the colorlist to predict the message content, i.e. majority voting. To attain a higher match rate at the expense of rate $E$, one can use error correction codes such as \cite{wicker1999reed} \textit{prior} to the encoding function and \textit{after} the decoding function.

\subsubsection{Message Correction with Feedback}
One key characteristic of our $p(y|x)$ is that we can instantly check whether the message was correctly transmitted by examining whether the sampled token is in the correct colorlist. This property resembles the settings of error correcting codes with feedback, in which the receiver can send feedback to the sender after receiving the message\citep{berlekamp1964block, gupta2023binary}. One can take advantage of this property by adapting the magnitude of the bias during encoding when the majority vote of a given position differs from the actual message. 

We provide some preliminary results of taking advantage of feedback during message encoding. One simple scheme is adapting the magnitude of the bias so that when the message is not correctly encoded, we enlarge the bias. Concretely, for $0 \leq t \leq T$ that is allocated to position $p$, if the current max colorlist does not match the actual message content, i.e. $\mathbf{m}[p] \neq \text{argmax}_j \mathbf{W}[j]$, we use a larger bias $\tilde \delta > \delta$. The results in Table \ref{tab:adaptive-bias} show that all lead to an increase in the multi-bit accuracy. However, we observed this came with a degradation in text quality measured by automatic metrics. We leave finding better methodology as a future work.

\subsection{Extending \ours~to other methods}\label{appendix:block-allocation}
\textbf{Block Allocation}
Instead of allocating a single position as done in \ours, we can allocate a block of message, after which techniques of \cs~can be used to encode the block message. This ensemble approach enables the prior works to embed longer messages. Deriving it name from Position Allocation, we dub this as Block Allocation.
\FrameSep1pt{
\begin{framed}
\noindent\textbf{\texttt{Block Allocation}}
\begin{enumerate}[noitemsep]
    \item Compute $s=f(X_{t-h:t-1})$.
    \item Chunk message in $n$ blocks. $\textbf{m}=[\textbf{m}_1, \dots, \textbf{m}_n]$ where $\textbf{m}_n\in\Sigma^{\frac{b}{n}}$
    \item $p \leftarrow \texttt{sample}([n])$ using $s$ as seed. 
    \item Run \cs~with message as $\textbf{m}_p$
\end{enumerate}
\end{framed}}

At decoding, we predict the message for each block and concatenate them. As a preliminary experiment, we use \texttt{Block Allocation} with \cs~ using $n$=4 blocks. \texttt{Block Allocation} can embed 24-bit messages with .901 bit accuracy (c.f. \cs~achieves .775) and 32-bit with .871 accuracy. 

\noindent\textbf{Extension to Other Zero-bit Watermarking}
\citet{openai-watermark} is another line of work in zero bit watermarking that modifies the sampling process by generating a secret vector $\mathbf{r}\in [0,1]^{|\mathcal{V}|}$ based on the random seed $s$. Given the original probability distribution $\mathbf{p}^{|\mathcal{V}|}$, the token with both large $p_v$ and $\mathbf{r}_v$ is favored by choosing  
\begin{equation}
    x = \text{argmax}_{v\in \mathcal{V}}\mathbf{r}_v^{\sfrac{1}{\mathbf{p}_v}}.
\end{equation}

We can adapt our position allocation method to this as well by preceding the above step with position allocation. Then, the secret key can be modified depending on the message content by the following rule:
\begin{equation}
    \mathbf{r} = 
        \begin{cases}
            \mathbf{r} & \text{if }\mathbf{m}[p] = 0  \\
            \mathbf{1} -\mathbf{r} & \text{if }\mathbf{m}[p] = 1 \\
        \end{cases}
\end{equation}
where $\mathbf{1}$ is a vector with 1 in all the elements. Analogous to favoring mutually exclusive colorlists, this allows favoring different tokens depending on the message content. At decoding time, we can similarly maintain a counter for each position for the two cases.

\subsection{Implementation, Hardware, Code Details}\label{appendix:imp}
We follow \cite{kirchenbauer2023watermark} in most experimental settings. 
For the hashing scheme in the main paper, we use LeftHash scheme with context window $h=1$. In the appendix, we provide results for the SelfHash scheme. For further discussions regarding the hash scheme see Appendix \ref{appendix:hash}.
To generate sequences with the desired token length $T$, we generate with the max token set as $T$. Then we filter out the watermarked and non-watermarked sequences with token lengths under $T_\text{low}=T-\tau$. We set $\tau$=25, except for the LFQA dataset, which was set to $\tau$=50 as it has instructions that state to generate answers with 200-300 words. For generation, we use sampling with a temperature of 0.7. For each bit-width, a new set of generations had to be made as the length of the message differed. 

For the copy-paste attack, we sample a random non-watermarked text and truncate to have the same length. Then, a position is randomly sampled to insert a $p$ percentage of the watermarked text into the non-watermarked text. We experiment with varying degrees of $p$ (10\%$\sim$ 50\%).

We used \texttt{float16} for all our models during generation. Our experiment was run on a single NVIDIA A100. For T=250, generating around 500 watermarked and non-watermarked samples took approximately 200 minutes for the left hash scheme. When using the self-hash scheme, this took significantly longer ($\sim$ 550 minutes). Our implementation is based on the official codebase of \citet{kirchenbauer2023watermark}: \url{https://github.com/jwkirchenbauer/lm-watermarking}. 

For baselines, we use the official repository of \citet{fernandez2023three}\footnote{\url{https://github.com/facebookresearch/three_bricks}} and \citet{wang2023towards}\footnote{\url{https://github.com/lancopku/codable-watermarking-for-llm}}. For \mh~, following the same configuration presented in their work (GPT-2 as the proxy model) cannot watermark the outputs of LLaMA-based models due to the difference in the tokenizers. Consequently, we resort to the Vanilla Marking scheme. This makes all the other factors equivalent for the three methods (\ours, \mh, \cs) except the message encoding function $\mathcal{E}$ described in \cref{sec:method}. Besides, we believe this has little to no effect on the watermark performance, since the use of proxy model is intended to enhance the quality of the text (in terms of perplexity) rather than the strength of the watermark.

\subsection{Metrics: Bit Accuracy, Text Quality}\label{appendix:bit-acc-as-metric}

\begin{figure}
    \includegraphics[width=.45\textwidth]{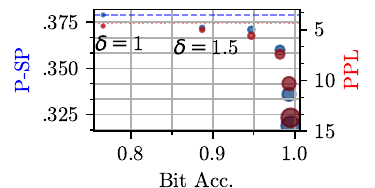}
    \caption{Text quality vs. $\delta$ across bias@T=100,b=8}
    \label{fig:delta}
\end{figure}

\textbf{Text Quality Metrics} 
We use the automatic metrics used in \citet{kirchenbauer2023reliability} such as perplexity (PPL) using a larger oracle model (LLaMA-2-13B) and semantic similarity based on a paraphraser model \citep[P-SP]{wieting2022paraphrastic}. Using P-SP, we measure the semantic similarity between the human text and watermarked text given the same prompt. While human evaluation is considered to be the golden label, our main purpose is to show that our multi-bit watermarking does not degrade the quality compared to zero-bit watermarking. Moreover, the effect of watermarking on the text quality \textit{compared to no watermarking} shows promising results in human evaluations when sufficiently large models are used for open-ended generation by \citealt{kirchenbauer2023reliability} (Appendix A.2 and A.9). Additionally, \citet{fernandez2023three} also demonstrate that watermarking does not lead to noticeable performance degradation even on benchmarks with non-ambiguous answers such as coding and math especially with sufficiently larger models, albeit at a small bias. We further show in Fig. \ref{fig:delta} the trade-off curve between bit accuracy and text quality. The size indicates the magnitude of bias ($\{$1, 1.5 2, 3, 4, 5$\}$) and horizontal dashed lines indicate non-watermarked counterparts. Analysis of text quality shows $\delta=2$ lies at a good trade-off point.

\noindent\textbf{Bit Accuracy for Multi-bit Watermark}
In our experiments, we used bit accuracy (error) as our metric for multi-bit watermark performance. This is a general metric that is independent of the downstream application or the encoding scheme.
However, computing the exact match of a message should be done dependent on the context.
To illustrate this, we start with some examples. First, consider the case where the encoding scheme to identify users is simply assigning a message to each user. Then, by embedding 4-bit message one can encode $2^{4}$ different users : $\mathbf{m}$=`0000' for Bob, $\mathbf{m}$=`0001' for Alice, and so on. For such a scenario, one might be interested in computing the exact match of the 4-bit message, also known as the packet error ratio.  While this encoding scheme enables tracing back to the exact users at low load capacity, this is extremely inflexible as it cannot handle influx or outflux of users.

Conversely, one can turn to a more flexible encoding scheme by encoding each character. Using UTF-8, this requires 8 bits per character, which would mean 40 bits is required just for encoding 5 character user ID. For this scenario, one might be more interested in computing the packet error ratio of each character or the entire 40-bit message. A more realistic encoding scheme will be somewhere between the middle, which uses a more efficient representation, e.g. by merging often-used bytes as done in Byte pair encoding \citep{gage1994new}. Added with error correction codes such as the Reed-Solomon code~\citep{wicker1999reed}, this allows a more robust representation. Since focusing on a single type of encoding scheme -- and more fundamentally, what information to embed -- narrows down the potential applications, we present bit accuracy in our main experiments as done in previous works in the literature~\citep{zhu2018hidden, luo2020distortion, yang2022tracing, yoo2023robust, fernandez2023stable}. For T=250, the packet error ratio for the 8-bit message was 7.1\%, which is +5.7 \% higher than the bit error rate. With 16-list decoding, this is reduced to 2.4\%.

Another metric considered in Table III of \citet{fernandez2023three} was combining the detection scheme and packet error ratio. In this scenario, they assume an encoding scheme of assigning each user to a single message and compute the percentage of finding the exact user given a fixed false positive rate. At FPR=$1e^\text{-3}$ and using 8-bit message (256 users), we can correctly identify 90.5\% cases. Our true positive rate was computed by the setting used in Table \ref{tab:tpr}.

\subsection{Discussion on the Hashing Scheme}\label{appendix:hash}
The hashing scheme for generating the seed plays a significant role in watermarking. For our MPAC, the hashing scheme is employed once for position allocation and once for permuting the vocabulary list. Here, we discuss some implications of the design choices. 

To recap, the function $f(X_{t-h:t-1})$ is used to hash $h$ most recent tokens before generating the $t^\text{th}$ token. Following the terminology of \citet{kirchenbauer2023reliability}, LeftHash takes the leftmost token, while SelfHash is determined in a slightly more complex way that is dependent on the $t^\text{th}$ token (see Algorithm 1 of \citet{kirchenbauer2023reliability}). The context width and the hashing scheme determine robustness and quality (diversity) trade-offs. For our experiments, we use the two configurations (LeftHash with $h$=1 and SelfHash with $h$=4) proposed in the previous work found to be effective in the two aspects without further fine-tuning. 

As expected by the trade-off, the perplexity was slightly higher for LeftHash compared to SelfHash (5.1 vs. 4.9 on average for 250 tokens), while P-SP was at the same level. One clear distinction between the two schemes was the encoding time latency. As SelfHash iteratively searches for tokens, this took significantly longer than the LeftHash scheme, which had nearly no overhead compared to no watermarking (\cref{appendix:imp} and Table \ref{tab:mic}). In addition, we observed that the sampled positions were not uniform for LeftHash with $h=1$ as shown in Tab. \ref{tab:hashing-scheme-pos-ratio} due to the reduced diversity of the tokens in the context width. Despite this, the multi-bit performance was similar for the two schemes (Table \ref{tab:fixedT} and \ref{tab:bpt}). A possible direction for improvement may be using different hashing schemes for position allocation (more robust) and vocabulary partitioning (more quality-focused).

\begin{table}[t]
    \centering
    \begin{tabular}{c|cccc}
        \toprule
        \multicolumn{5}{c}{\textbf{Ratio Sampled Position (Sorted)}} \\ 
        LeftHash ($h$=1) & 0.319	& 0.251 & 0.235 & 0.195 \\
        SelfHash ($h$=4) & 0.264    & 0.257 & 0.242 & 0.238  \\ 
        \bottomrule
    \end{tabular}
    \caption{Ratio of the sampled position for $b$=8,$r$=4 (four positions total) for the two hashing schemes for position allocation. 
}  \label{tab:hashing-scheme-pos-ratio}
\end{table}

\begin{table*}[t]
    \centering
    \begin{tabular}{c|cccc}
        \toprule
        \multicolumn{5}{c}{\textbf{Bit Acc. after Paraphrasing with DIPPER}} \\ 
        \multicolumn{1}{c}{Bit-width}  & 8 & 16 & 24 & 32 \\ \hline
        Best Prediction   &  .922 (.13) & .825 (.12) & .778 (.12) & .736 (.10) \\
        16-List Decoded   &  .982 (.05) & .924 (.08) & .864 (.10) & .801 (.09) \\
        \bottomrule
    \end{tabular}
    \caption{Robustness under paraphrasing using DIPPER (Lexical diveristy=20)}  \label{tab:dipper}
\end{table*}

\begin{table*}[t]
    \centering
    \begin{tabular}{c|c|cccc}
        \toprule
        &  & \multicolumn{4}{c}{DIPPER} \\
        & GPT-3.5 & Lex.=20 & Lex.=40 & Lex.=60 & \makecell{Lex.=60\\Ordering=60} \\
        \toprule
        P-SP  &  .815	&.933	&.897	& .844	&.827 \\
        \small Absolute Change in $\#$ of Words  & 36 & 13 & 16 & 19 & 20\\
        Bit Acc. & .733 &	.922 &	.849 &	.757 &	.719 \\
        \bottomrule
    \end{tabular}
    \caption{Comparison of the two paraphrasing method on text quality.}  \label{tab:dipper-semantic}
\end{table*}

\begin{figure*}
    \centering
    \includegraphics{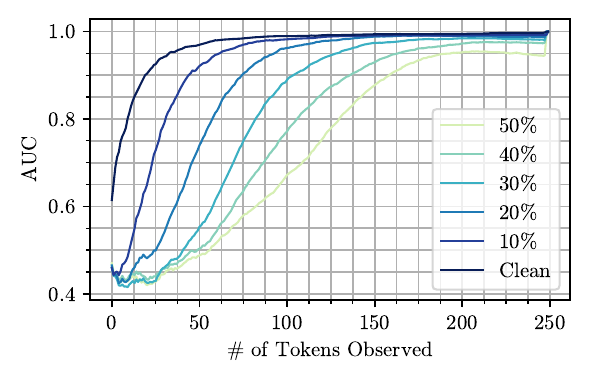}
    \caption{AUC vs. number of tokens observed when corrupted with copy-paste attack for 8-bit message.}
    \label{fig:auc-at-t-cp}    
\end{figure*}

\begin{figure*}
\begin{minipage}[htbp]{0.79\textwidth}
    \centering
    \includegraphics[width=\textwidth]{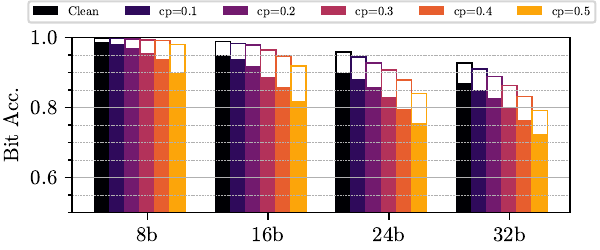}
    \vspace{-8mm}
    \subcaption{}\label{fig:robust-cp}
\end{minipage}
\hfill
\begin{minipage}[htbp]{0.2\textwidth}
    \centering
    \includegraphics[width=\textwidth]{fig/robustness-gpt.pdf}
    \vspace{-8mm}
    \subcaption{}\label{fig:robust-gpt}   
\end{minipage}
\vspace{-4mm}
\caption{Corrupted bit accuracy for (a) copy-paste attack controlled by the human text percentage at T-250 and (b) paraphrasing attack using GPT-3.5 embedding 8-bit messages at varying token lengths. For (b), we show multiple sizes of list ($|L|\in$\{2, 4, 8, 16\}) by color gradation as 8-bit has relatively small output space.}
\label{fig:robust-appendix}
\vspace{-4mm}
\end{figure*}

\begin{figure*}
    \centering
    \includegraphics{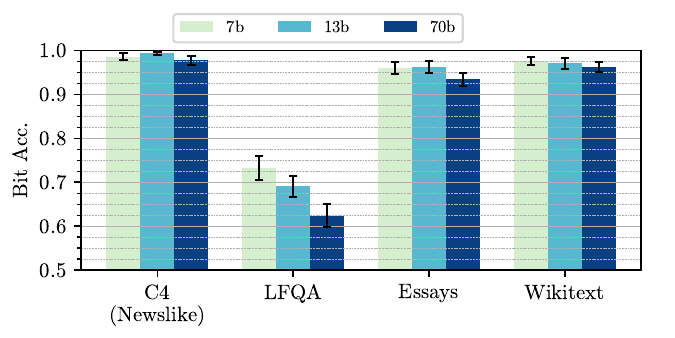}
    \caption{Multi-bit performance across datasets and model sizes.}
    \label{fig:model-dataset-ablation}
\end{figure*}

\begin{table}[t]
    \centering
    \begin{tabular}{c|c}
        \toprule
        \multicolumn{2}{c}{\textbf{Robustness on GPT-3.5 paraphrasing}} \\ 
        Methods & Bit Acc. \\
        \toprule 
        \ours & .733 (.19)	\\
        \cs$^{*}$  &  .655 (.25)   \\ 
        \mh$^{*}$  &  .540 (.24) \\
        \bottomrule
    \end{tabular}
    \caption{Bit accuracy after paraphrasing on $T$=250, $b$=8. For~\cs and~\mh, we use 100 samples due to resource constraint on GPT-3.5.}  
    \label{tab:gpt-paraphrasing}
\end{table}

\subsection{More on Robustness: Other Attacks, Detection}\label{appendix:dipper} 
We also test our watermark against DIPPER \citep{krishna2023paraphrasing}, which is a specialized paraphrasing model. DIPPER is parameterized by two scalers, which control lexical diversity and token order diversity. We first present the results across bit-width with a lexical diversity of 20 (out of 100). We see that the watermark fares considerably better than using GPT-3 attack in Table \ref{tab:dipper}.  

To see the magnitude of semantic drift of the two paraphrasing methods, we compute the P-SP between the original watermarked text and its paraphrased counterpart. We also compute the absolute change in the number of words. Table \ref{tab:dipper-semantic} demonstrates that paraphrasing using GPT-3.5 changes the semantic and the number of words greater than the setting used in Table \ref{tab:dipper}, which may explain why the multi-bit watermark performance is lower for GPT-3.5. When we control the diversity parameters of DIPPER, this is able to degrade the watermark performance as well as GPT-3.5.

Some other forms of possible attacks considered in the literature are word substitution, insertion, and deletion. Word substition is very similar to the copy-paste attack considered in the main paper. Our watermark scheme is also robust to partial insertion and deletion of words as MPAC relies on the local context to synchronize the positions of the message and the ordering of the vocabulary. 

\noindent\textbf{Robustness of zero-bit Watermark} Here we provide results for the detection performance under corrptuion. We use the copy-paste attack with the attack percentage ranges of \{10\%, 20\%, 30\%, 40\%, 50\%\} and compare the AUC vs. number of tokens observed curve similar to Fig. \ref{fig:auc-at-t-cp}. While the detectability is noticeably affected, the final AUC is recovered to a large degree only after observing 250 tokens. In order of the attack strength, the final AUC's are .992, .987, ,980, ,971, .942, respectively.  For the zero-bit counterpart, all the scores are over .990.


\subsection{Ablations on Datasets and Model Sizes}\label{appendix:models} 
We show additional results on other datasets and model sizes in Fig. \ref{fig:model-dataset-ablation}. C4 news-like subset is the dataset we used for our main experiment. "Long-form Question-Answering" (LFQA) is a dataset curated by \citet{krishna2023paraphrasing} on the Reddit’s “Explain Like I’m Five” (ELI5) forum. The Essays dataset comprises pairs of instructions and essays~\citep{essay}. Wikitext~\citep{merity2016pointer} comprises Wikipedia article. We use the `wikitext-2' subset. For LFQA, we use the finetuned version, LLaMA-2-Chat, specialized for chats as they explicitly have questions or instructions as prompts.

It is apparent that the watermark performance is affected by the text distribution. When the entropy of the vocabulary distribution is low (low diversity), there is little room for encoding the message with a fixed bias, which has been observed in zero-bit watermarking as well where the watermark performance suffers for low entropy text distributions such as coding~\citep{lee2023wrote, kirchenbauer2023reliability}. For our multi-bit case, this means the load capacity is inherently low for such text distributions. This is especially observed for LFQA, in which the model consistently starts the response by restating the question (e.g. \textit{"The reason for [Question] is $\dots$"}). Across the model scale, the trend is not as apparent although we found that the largest model consistently has a lower performance. This hints that the entropy of the vocabulary distribution is lower for the largest model, which might explain the higher text quality in general when we increase the model size. Larger models might have the capacity to form high-quality sequences even when the text distribution is altered by increasing the entropy via temperature or explicitly increasing the magnitude of the bias during watermarking. We leave this as a future work.

\begin{table}[t]
    \centering
    \small
    \begin{tabular}{c|ccccc}
        \toprule
        \multicolumn{6}{c}{\textbf{True Positive Rate}} \\ 
        \multicolumn{1}{c}{Bit-width} &  0 & 8 & 16 & 24 & 32 \\ \hline
        FPR=$1e^\text{-2}$ & 0.999&  0.986&  0.974&  0.964 & 0.958 \\
        FPR=$1e^\text{-3}$ & 0.997 &0.974 &0.956 &0.943 &0.915 \\ 
        FPR=$1e^\text{-4}$ & 0.997& 0.96& 0.934& 0.905& 0.88 \\ 
        FPR=$1e^\text{-5}$ & 0.994 &0.951 &0.907 &0.851 &0.793 \\ 
        \bottomrule
    \end{tabular}
    \caption{True positive rate at a fixed false positive rate across bit-widths. We use $\sim 500$ positive sample and $\sim$100,000 negative samples. We only count the unique tokens following \citep{kirchenbauer2023watermark, fernandez2023three}. This has an effect of removing outlier human text samples that have exceptionally high scores.
}  \label{tab:tpr}
\end{table}

\subsection{Tabular Results}\label{appendix:tables}
Here we present the numerical results of the experiments done in the main paper. Numbers in the parenthesis signify the standard deviation. 
\begin{itemize}
    \item Table \ref{tab:comparison-with-others} and \ref{tab:comparison-with-others2} $\leftrightarrow$ Figure \ref{fig:main-comparison} show the comparisons with baseline methods.
    \item Table \ref{tab:delta} $\leftrightarrow$ Figure \ref{fig:delta} show the relationship between $\delta$ vs. text quality and watermark strength.
    \item Table \ref{tab:clean-radix} $\leftrightarrow$ Figure \ref{fig:ours-only} left compare the different configurations of radix and colorlist proportion. 
    \item Table \ref{tab:fixedT} $\leftrightarrow$ Figure \ref{fig:ours-only} left show the multibit watermark performance on a fixed token length.
    \item Table \ref{tab:bpt} $\leftrightarrow$ Figure \ref{fig:ours-only} right show the multibit watermark performance on a fixed load capacity (bits per token). 
    \item Table \ref{tab:robustness-cp} $\leftrightarrow$ Figure \ref{fig:robust}a show the multibit watermark performance under copy-paste corruption. 
    \item Table \ref{tab:robustness-gpt} $\leftrightarrow$ Figure \ref{fig:robust}b show the multibit watermark performance under paraphrasing. 
\end{itemize}

\begin{table*}[t]
\centering
\begin{adjustbox}{width=.5\textwidth}
\begin{tabular}{c|cccc}
    \toprule
     & \multicolumn{4}{c}{\textbf{B=8,T=250}} \\ 
    \toprule
     Copy-Paste ($p$) &  Clean  &  cp=10\% &  cp=30\%  &  cp=50\% \\
    \toprule
    Ours& .986 (.06) & .981 (.07) & .956 (.10)  & .900 (.13)  \\
    FCT+EMS & .979 (.10)  & .943 (.17) & .858 (.24) & .800 (.28)     \\
    FCT+Greenlist& .995 (.05)  & .988 (.08) & .970 (.12) & .908 (.20) \\
    CTWL & .977 (.11) & .973 (.12) & .951(.16) & .858(.24)  \\
    \bottomrule
\end{tabular}
\end{adjustbox}
\caption{Comparison of multibit watermark performance with other methods on clean and corrupted settings. For corruption, we use the copy-paste attack. *The load capacity of FCT+Greenlist is limited to 15-bit.}\label{tab:comparison-with-others}
\end{table*}

\begin{table*}[t]
\centering
\begin{adjustbox}{width=\textwidth}
\begin{tabular}{c|cccc|cccc}
    \toprule
     & \multicolumn{4}{c}{\textbf{B=16,T=250}} & \multicolumn{4}{c}{\textbf{B=24,T=250}}\\ 
    \toprule
     Copy-Paste ($p$) &  Clean  &  cp=10\% &  cp=30\%  &  cp=50\% & Clean  &  cp=10\% &  cp=30\%  &  cp=50\% \\
    \toprule
    Ours& .951 (.07) & .939 (.08) & .887 (.09)  & .819 (.12) & .899 (.09)&	.882 (.09)&	.830 (.10)&	.755 (.11) \\
    FCT+EMS &.905 (.20)  & .811 (.26) & .702 (.26) & .601 (.23) & .775 (.26)&	.729 (.24)&	.633 (.23)	&.513 (.13)   \\
    CTWL & ..936 (.18)	 &.909 (.20) &	.810 (.26)	 &.614 (.22) & .876 (.22)&	.828 (.25)&	.663 (.26)&	.516 (16)\\
    \bottomrule
\end{tabular}
\end{adjustbox}
\caption{Comparison of multibit watermark performance with other methods on clean and corrupted settings.}\label{tab:comparison-with-others2}
\end{table*}

\begin{table*}[ht]
\small
\centering
\begin{tabular}{c|ccccccc}
\toprule
$\delta$ & 0.5 & 1 & 1.5 & 2 & 3  & 4 & 5 \\\hline
Bit Acc.  & .626 (.19) & .766 (.18) & .887 (.15) & .947 (.11) & .982 (.08) & .993 (.05  & .995 (.05)  \\ 
P-SP (w/ reference) & .385 (.15) & .379 (.15) & .372 (.15) & .371 (.15  & .360 (.14) & .336 (.13) & .319 (.13)  \\
P-SP (w/ non-wm.) & .526 (.18) & .460 (.16) & .433 (.15) & .417 (.15) & .388 (.14) & .349 (.14) & .330 (.13)  \\
PPL & 4.41 (1.5) & 4.64 (1.8) & 5.01 (2.0) & 5.6 (2.0)  & 7.41 (2.7) & 10.3 (4.1) & 13.67 (5.9) \\ 
\bottomrule
\end{tabular}
\caption{Bit accuracy and text quality on embedding 8 bit-width message on T=250 across various magnitudes of bias $\delta$.} \label{tab:delta}
\end{table*}

\begin{table*}[ht]
\centering
\begin{tabular}{c|cccc}
    \toprule
    \multicolumn{5}{c}{\textbf{Bit Accuracy @ T=250}} \\ 
    \toprule
    Bit &  8  &  16 &  24  &  32 \\
    \toprule
    $\gamma$=.25,$r$=4& .986 (.06) & .951 (.07) & .900 (.09)  & .871 (0.08) \\
    $\gamma$=.25,$r$=2& .966 (.07)  & .905 (.08) & .858 (.08) & 0.820 (.08) \\
    $\gamma$=.50,$r$=2 & .978 (.05)  & .922 (.07) & .875 (.08) & 0.849 (.07)   \\
    \bottomrule
\end{tabular}
\caption{Multibit watermark performance measured by bit accuracy for varying configurations of colorlist proportion and radix.}\label{tab:clean-radix}

\end{table*}

\begin{table*}[ht!]
\centering
\begin{tabular}{c|cccc}
    \toprule
    \multicolumn{5}{c}{\textbf{Bit Acc. @ T=250}} \\ 
    \toprule
    Bit &  8  &  16 &  24  &  32 \\
    \toprule
    LeftHash($h=1$) & .986 (0.06) & .951 (.07) & .900 (.09) & .871 (0.08)   \\
    SelfHash($h=4$) & .976 (.08)  & .905 (.08) & .895 (.09) & .862 (.09)    \\ 
    \bottomrule
\end{tabular}
\caption{Bit accuracy for two different hash schemes for a fixed token length.}\label{tab:fixedT}

\begin{tabular}{c|cccccc}
    \toprule
    \multicolumn{6}{c}{\textbf{Bit Acc. @ BPT=.064}} \\ 
    \toprule
    T &  63  & 125 &  250  &  500 &  1000 \\
    Bit & 4 &  8 & 16  & 32 & 64 \\
    \toprule
   LeftHash($h=1$) & .961 (.13)  & .958 (.09)   & .951 (.07)   & .913 (.08) & .846 (.09) \\
   SelfHash($h=4$) & .952 (.13)  & .953 (.10) & .945 (.08) & .911 (.08)  & .850 (.08) \\
    \bottomrule
\end{tabular}
\caption{Bit accuracy for two different hash schemes for a fixed bits per token.}
\label{tab:bpt}
\vspace{-1mm}
\end{table*}

\begin{table*}[ht!]
\centering
\begin{tabular}{cc|cccccc}
    \toprule
    \multicolumn{8}{c}{\textbf{Copy-paste Attack}} \\ 
    \multicolumn{2}{c}{Attack Strength} &  Clean  &  10\% &  20\%  &  30\% &  40\% & 50\% \\ \hline
    \multirow{2}{*}{8-bit} & Best & .986 (.06)  & .981 (.07) & 0.971 (.08) & .956 (.10) & .938 (.12) & .900 (.13) \\
     & +16-List & .997 (.02)  & .997 (.02) & .995 (.03)  & .993 (.03) & .991 (.04) & .980 (.05) \\
     \hline 
    \multirow{2}{*}{16-bit} & Best & .951 (.07)  & .939 (.08) & .918 (.09)  & .887 (.09) & .858 (.11) & .819 (.12) \\
     & +16-List & .988 (0.04) & .983 (.04) & .978 (.05)  & .964 (.06) & .947 (.07) & .918 (.08) \\
     \hline
    \multirow{2}{*}{24-bit} & Best & .899 (.09)  & .882 (.09) & .858 (.10)  & .830 (.10) & .797 (.11) & .755 (.11) \\
     & +16-List & .959 (.06)  & .944 (.06) & .927 (.08)  & .907 (.08) & .879 (.09) & .840 (.09) \\
     \hline
    \multirow{2}{*}{32-bit} & Best & .871 (.08)  & .851 (.09) & .828 (.09)  & .801 (.09) & .765 (.09) & .723 (.1)  \\
     & +16-List & .927 (.07)  & .910 (.08) & .888 (.08)  & .863 (.08) & .831 (.09) & .792 (.09) \\
    \bottomrule
\end{tabular}
\caption{Robustness when certain percentage of human text is mixed into the watermarked text.} \label{tab:robustness-cp}
\end{table*}

\begin{table*}[h!]
\centering
\begin{adjustbox}{width=0.5\textwidth}
\begin{tabular}{cc|ccc}
    \toprule
    \multicolumn{5}{c}{\textbf{GPT-3.5 Paraphrasing}} \\ 
    \multicolumn{2}{c}{Token Length} &  250T  &  400T &  500T  \\ \hline
    \multirow{5}{*}{8-bit} 
     & Best & .733 (.19)  & .792 (.19) & .795 (.19)  \\
     & +2-List & .825 (.16)  & .874 (.15) & .875 (.15)   \\
     & +4-List & .856 (.14)  & .894 (.13) & .898 (.13)   \\
     & +8-List & .893 (.12)  & .924 (.11) & .928 (.11)   \\
     & +16-List & .911 (.10)  & .934 (.09) & .939 (.09)   \\
    \bottomrule
\end{tabular}
\end{adjustbox}

\begin{tabular}{p{0.6\textwidth}}
    \textbf{Prompt} \\ 
    “As an expert copy-editor, please rewrite the following text in your own voice while ensuring that the final output contains the same information as the original text and has roughly the same length. Please paraphrase all sentences and do not omit any crucial details. Additionally, please take care to provide any relevant information about public figures, organizations, or other entities mentioned in the text to avoid any potential misunderstandings or biases.” \\
    \bottomrule
\end{tabular}
\caption{Robustness when paraphrased with GPT-3.5.}\label{tab:robustness-gpt}
\end{table*}

\subsection{Generation Samples}\label{appendix:generation samples}
We show below in Table \ref{table:samples} generated samples.

\begin{table*}[h!]
\centering
\small
\renewcommand*{\arraystretch}{2.0}
\begin{tabular}{p{.3\textwidth}|p{.3\textwidth}|p{.3\textwidth}}
\toprule
\multicolumn{1}{c}{Prompt} & \multicolumn{1}{c}{Non-Watermarked} & \multicolumn{1}{c}{Watermarked}  \\\midrule
A blind man in his 60s is searching for the young man who pulled him back from an approaching train and saved his life last Thursday, Nov. 9.
Mike Wyatt stood at the stairs to the Peoria train station in Aurora, Colorado, on Monday, looking for the young man who saved his life days earlier 
& 
``He told me he was 20 years old and he was going to college," Wyatt told reporters. ``He said he was going to be a doctor."
Wyatt, who is blind and has been for most of his life, was at the train station ready to catch a train to the airport when he slipped and almost fell in front of an oncoming train.[...]
& 
and has given him new life.
Inside the station, he spoke to the counter clerk, hoping the clerk might have information that might lead him to the young man, who jumped at the chance to help Wyatt when he saw he was in danger. But, he said, he found nothing.[...]
\\\hline
[...]
Shortly after the release of the lead single “Life’s About to Get Good,” it hit number one on the Billboard 200. Of course, fans were simply dying to see how their favourite country icon was doing, so
 &
 the album was released on CD, vinyl, and digital. But, Shania took things to the next level and released a deluxe version with added tracks. [...]
 &
 the song was released a few months earlier in June.
``OK so I am sooooo proud that my song Life's About to Get Good is now the number one song on Billboard country,” she tweeted back in June.[...]
 \\\hline
 [..] Amazon’s offices will be located within an already-established special tax district where a portion of the property tax revenue goes toward infrastructure improvements such as parks and wider sidewalks.
The incentive agreement says that half of any new revenue from that district starting in 202
 &
 3 would go to Amazon to help pay for employee benefits and transportation, but the county has said it would not be required to pay those benefits.[...]
 &
 7 would go to Amazon through a 15-year payment schedule. This amount would grow to more than 60 percent if the company exceeds 10.24 million square feet of office space. Amazon is expected to reach that level in 2027.[...] \\\hline

[...]
"Below the surface of the Sun a dynamo process is working creating magnetic field," Harra explained in an email interview. "When this becomes buoyant it can rise to the surface of the Sun,
&
and this is what we call a sunspot."
The new research shows that the initial explosion was actually a series of smaller blasts that combined to create the X-7 flare. 
&
and form coronal holes and sunspots. This happens in a fairly homogeneous area on the Sun, so there can be several sunspots with a single magnetic field underneath. \\\hline

[...]The merge listing the most important changes to Linux 3.8's sound subsystem includes some other changes to audio drivers.
The kernel now includes a driver for human interface devices (HIDs) that use I2C (1, 2 and others), using the "HID over I2C" protocol designed by Microsoft and implemented in WindowsÂ 
&
 7 and later versions of the operating system.
The kernel now has a driver for the Samsung Galaxy S III smartphone's touchscreen (1, 2 and others), and the rt2800usb driver, for the RaLink RT2800USB WLAN chip, now supports devices that have Bluetooth 3.0 (1, 2).[...]
&
 7 and Windows Vista. The drivers can read out data from HIDs and set the appropriate commands to them. An example of such a device is a BT-USB adapter.
The sound subsystem now supports two new, high-quality audio codecs (1, 2):[...] \\ 
\bottomrule
\end{tabular}
\vspace{1em}
\caption{Randomly sampled examples of watermarked texts on the C4 newslike subset with 100\% bit accuracy. Samples are truncated for readability.}
\label{table:samples}
\end{table*}

\end{document}